\documentclass[final]{cvpr}
\usepackage{amsmath,amsfonts,bm}

\def\eqref#1{equation~\ref{#1}}
\def\1{\bm{1}}

\def\vw{{\bm{w}}}

\DeclareMathAlphabet{\mathsfit}{\encodingdefault}{\sfdefault}{m}{sl}
\SetMathAlphabet{\mathsfit}{bold}{\encodingdefault}{\sfdefault}{bx}{n}

\DeclareMathOperator*{\argmax}{arg\,max}

\usepackage{times}
\usepackage{epsfig}
\usepackage{graphicx}
\usepackage{amsmath}
\usepackage{amssymb}

\usepackage{url}
\usepackage{subcaption}
\usepackage{graphicx}
\usepackage{multirow}
\usepackage{booktabs}
\usepackage{color}
\usepackage{xcolor}
\usepackage{wrapfig}

\definecolor{bostonuniversityred}{rgb}{0.8, 0.0, 0.0}

\newif\ifshow
\showtrue
\ifshow
\newcommand{\ari}[1]{\textcolor{blue}{[\textbf{Ari:} #1]}}
\newcommand{\rudy}[1]{\textcolor{bostonuniversityred}{#1}}

\newcommand{\todo}[1]{\textcolor{red}{[\textbf{TO DO}: #1]}}
\else
\newcommand{\ari}[1]{}
\newcommand{\rudy}[1]{}
\newcommand{\todo}[1]{}
\fi

\usepackage[pagebackref=true,breaklinks=true,colorlinks,bookmarks=false]{hyperref}

\def\cvprPaperID{2} % *** Enter the CVPR Paper ID here
\def\confYear{CVPR 2021}

\begin{document}

%%%%%%%%% TITLE
\title{Width transfer: on the (in)variance of width optimization}

\author{Ting-Wu Chi$\text{n}^1$\thanks{This work is mostly done when Ting-Wu works as a research intern at Facebook AI Research.}, Diana Marculesc$\text{u}^{12}$, Ari S. Morco$\text{s}^3$\\
Carnegie Mellon Universit$\text{y}^1$, The University of Texas at Austi$\text{n}^2$, Facebook AI Researc$\text{h}^3$\\
{\tt\small tingwuc@andrew.cmu.edu, dianam@utexas.edu, arimorcos@fb.com}}
\maketitle

%%%%%%%%% ABSTRACT
\begin{abstract}
    Optimizing the channel counts for different layers of a CNN has shown great promise in improving the efficiency of CNNs at test-time. However, these methods often introduce large computational overhead (\textit{e.g.}, an additional $2\times$ FLOPs of standard training). Minimizing this overhead could therefore significantly speed up training. In this work, we propose width transfer, a technique that harnesses the assumptions that the optimized widths (or channel counts) are regular across sizes and depths. We show that width transfer works well across various width optimization algorithms and networks. Specifically, we can achieve up to $320\times$ reduction in width optimization overhead without compromising the top-1 accuracy on ImageNet, making the additional cost of width optimization negligible relative to initial training. Our findings not only suggest an efficient way to conduct width optimization, but also highlight that the widths that lead to better accuracy are invariant to various aspects of network architectures and training data.
\end{abstract}

\section{Introduction}\label{sec:intro}
Better designs for the number of channels for each layer of a convolutional neural network (CNN) can lead to improved test performance for image classification without requiring additional floating-point operations (FLOPs) during the forward pass at test time. Specifically, by optimizing the channel widths, improvements of up to 2\% top-1 accuracy for image classification on ImageNet can be achieved without additional FLOPs~\cite{guo2020dmcp,gordon2018morphnet,yu2019autoslim,chin2020pareco}. However, designing the layer by layer width multipliers for efficient CNNs is a non-trivial task that often requires intuition and domain expertise together with trial-and-error to do well. To alleviate the labor-intensive trial-and-error procedure, width optimization algorithms have been proposed ~\cite{liu2019metapruning,he2018amc,guo2020dmcp,dong2019network,gordon2018morphnet} to automatically determine the width of a convolutional neural network. A width optimization algorithm takes as input an initial network and a training dataset, and outputs a set of optimized widths for each layer. When these optimized widths are applied to the initial network and the network trained from scratch, one can achieve better validation accuracy compared to training a network of the original widths without incurring additional test-time FLOPs. Such algorithms can be seen as neural architecture search algorithms that search for layer-wise channel counts to maximize validation accuracy subject to test-time FLOPs constraints. In contrast to channel pruning, which seeks to reduce FLOPs without reducing accuracy, width optimization instead aims to improve the accuracy while maintaining the same number of FLOPs.
%designing the width for each layer based on computational methods has received growing interests. Examples include using reinforcement learning~\cite{he2018amc}, evolutionary algorithms~\cite{liu2019metapruning,chin2020towards}, and differentiable parameterization~\cite{guo2020dmcp,dong2019network,ning2020dsa} to optimize for layer widths. \ari{Should add here a clear contrast with pruning}

However, these methods often add a large computational overhead necessary for the width optimization procedure. Concretely, even for efficient methods that use differentiable parameterization~\cite{guo2020dmcp}, width optimization takes an additional $2\times$ the training time. To contextualize this overhead, using distributed training on 8 V100 GPUs, it takes approximately 100 GPU hours to train a ResNet50 on the ImageNet dataset~\cite{radosavovic2020designing}. Including the width optimization overhead, it therefore takes 300 GPU hours for both width optimization using differentiable methods~\cite{guo2020dmcp} and training the optimized ResNet50. Additionally, width optimization algorithms are often parameterized by some target test-time resource constraints, \textit{e.g.}, FLOPs. As a result, the computational overhead scales linearly with the number of target constraint levels considered, which can be exceedingly time-consuming for optimizing CNNs for embodied AI applications~\cite{chin2020towards}. Reducing the overhead for width optimization, therefore, would have material practical benefits.

%If \rudy{similar} $\mathcal{C}$s share similar $w^*$s structurally, we can downsize $\mathcal{C}$ to reduce the computational overhead incurred by $\mathcal{A}$ as shown schematically in Figure~\ref{fig:flow}.

\begin{figure*}[t]
    \centering
    \includegraphics[width=0.7\textwidth]{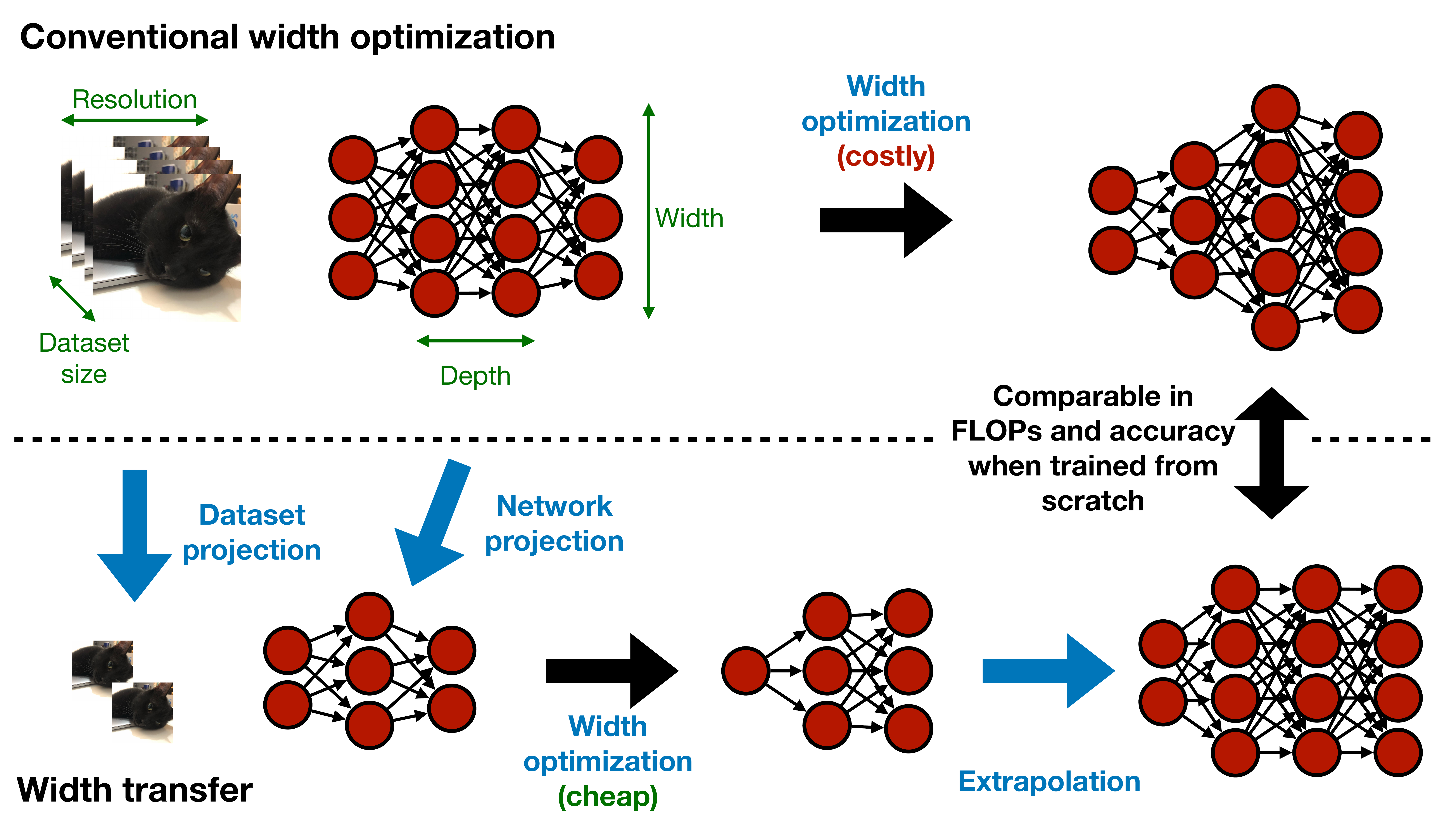}
    \caption{The top row shows the conventional width optimization approach, which takes a training dataset and a seed network, and outputs a network with optimized widths. The bottom row depicts our idea of width transfer, where width optimization operates on the down-scaled dataset and seed network. We then use a simple function to extrapolate the optimized architecture to match the original network. Compared to direct width optimization, our empirical findings suggest that width transfer has similar performance, but has the benefit of drastically lower overhead.}
    \label{fig:flow}
\end{figure*}

Fundamentally, one of the key reasons why width optimization is so costly is due to its limited understanding by the community. Without assuming or understanding the structure of the problem, the best hope is to conduct black-box optimization whenever training configurations, datasets, or architectures are changed. In this work, we take the first step to empirically understand the structure underlying the width optimization problem by changing network architectures and the properties of training datasets, and observing how they affect width optimization. 
% Such sensitivity analysis techniques have been used in various contexts in the deep learning literature for empirically unveiling the black-box of deep learning~\cite{morcos2018insights,tan2019efficientnet,li2018visualizing}. Similarly, we manipulate the network architectures and dataset properties to aid in our understanding of the (in)variances of width optimization. To the best of our knowledge, despite the community's great interests in channel optimization~\cite{deng2020model}, there is no systematic study on how the dataset and network affect channel optimization. However, such a study is critical for understanding and building better width optimization methods.

If similar inputs to the width optimization algorithms result in similar outputs, one can exploit this commonality to reduce the width optimization overhead, especially if the two input configurations have markedly different FLOPs requirements. As a concrete example, if optimizing the widths of a wide CNN (high FLOPS) and a narrow CNN (low FLOPs) results in widths that differ only by a multiplier, one can reduce the computational overhead of width optimization by computing widths for the low FLOPS, narrow CNN and adjusting them to accommodate the high FLOPs, wide CNN. To study the aforementioned commonality, we propose \textit{width transfer}, a novel paradigm for efficient width optimization. In width transfer, one first projects the network and the dataset to their smaller counterparts, then one executes width optimization with the smaller network and dataset, and finally one extrapolates the optimized result back as shown schematically in Figure~\ref{fig:flow}.

In other words, we would like to understand if the following invariances hold for optimized widths:
% \ari{I don't love the A1, A2 etc. labels, though it's nice to refer to it in the next paragraph. What does A mean? What about "Invariance 1: sample size", "Invariance 2: spatial resolution" etc.? Def open to other ideas if you have them}
\begin{enumerate}
    \item \textbf{Sample size}: The optimized widths are minimally affected by the size of the dataset when the dataset's distribution is approximately identical (\textit{i.e.}, uniform down-sampling in a class-balanced fashion).
    \item \textbf{Spatial resolution}: The optimized widths are merely affected by the image resolutions.
    \item \textbf{Channel magnitude}: The ratios between the optimized widths and the un-optimized ones are roughly constant regardless of the absolute channel counts of the un-optimized network.
    \item \textbf{Within-stage channel counts}: The optimized widths are similar when they belong to the same stage of a network where stage is defined by grouping the blocks with the same input resolution~\cite{he2016deep}.
\end{enumerate}

% However, the idea of width transfer presents challenges if the network projection results in a shallower network since the optimized network has fewer layers compared to its pre-projection counterpart. To tackle this challenge, we propose two layer-stacking strategies for matching their layer dimensions for a successful width transfer.
Based on a comprehensive empirical analysis, we provide the following contributions:
% In addition to the potential efficiency benefits from understanding the structure of the width optimization problem, such an exploration can also have scientific benefits. Specifically, via sensitivity analysis, we can provide quantitative results to the following question: Do the depth and width of a network, number of training samples, and the resolution of input images affect width optimization? And if so, by how much? Based on a comprehensive empirical analysis, we provide the following contributions:
\begin{itemize}
    \item We propose \textit{width transfer}, a novel paradigm for efficient width optimization. Additionally, we propose two novel layer-stacking methods to transfer width across networks with different layer counts. %\ari{I like this but the challenge of extrapolating width multipliers across depth isn't present anywhere in the intro. Can we add a comment to this effect?}
    \item We find that the optimized widths are highly transferable across network's initial width and depth, and across dataset's sample size and resolution.
    \item We demonstrate a practical implication of the previous finding by showing that one can achieve $320\times$ reduction in width optimization overhead for a scaled-up MobileNetV2 and ResNet18 on ImageNet with similar accuracy improvements, effectively making the cost of width optimization negligible relative to initial model training.
    \item With controlled hyperparameters over multiple random seeds on a large-scale image classification dataset, we verify the effectiveness of width optimization methods proposed in prior art. This is in contrast with prior work which borrows numbers from other papers that might not have the same training hyperparameters. However, we also find that, for a deeper ResNet on ImageNet, width optimization has limited benefits (Fig.~\ref{res18-depth-aavg}).
\end{itemize}

\section{Width optimization}
The layer-by-layer widths of a deep CNN are often regarded as a hyperparameter optimized to improve classification accuracy. Specifically, the width multiplier method~\cite{howard2017mobilenets} was introduced in MobileNet to arrive at models with different FLOPs and accuracy profiles and has been widely adopted in many papers~\cite{he2018soft,tan2019efficientnet,chin2020one,ding2019regularizing}.

Besides simply scaling the width uniformly for all the layers to arrive at CNNs with different FLOPs, optimizing the width for each layer has received growing interest recently as a means to improve the efficiency of deployed deep CNNs. A width optimization algorithm, $\mathcal{A}$, takes in a training configuration, $\mathcal{C}=(\mathcal{D}, \mathcal{N})$, and outputs a set of optimized widths, $\vw^*$. $\mathcal{C}$ consists of initial network, $\mathcal{N}$, and training dataset, $\mathcal{D}$. Concretely, the goal of $\mathcal{A}$ is to solve the following optimization problem:
\begin{align}\label{eq:width_opt}
    \begin{split}
        \vw^*=\argmax_{\vw} Acc_{\text{val}}\left(\theta(\mathcal{N}\times \vw,\mathcal{D}), \mathcal{N}\times \vw\right)\\
        \text{s.t.}~\text{FLOPs}(\mathcal{N})=\text{FLOPs}(\mathcal{N}\times \vw),
    \end{split}
\end{align}

where $\vw$ is a width vector with $L$ dimensions, where $L$ is the number of convolutional layers. $\mathcal{N}\times \vw$ is applying width $\vw$ to a network $\mathcal{N}$ by scaling the number of channels for layer $i$ from $F_i$ to $w_iF_i$. $\theta()$ is the standard training procedure that takes in a CNN and a training dataset and outputs trained weights, \textit{e.g.}, stochastic gradient descent with a fixed training epoch. Lastly, $Acc_{\text{val}}$ is a function that evaluates the validation accuracy given the trained weights and the CNN architecture.

\begin{figure}[h]
    \centering
    \includegraphics[width=0.3\textwidth]{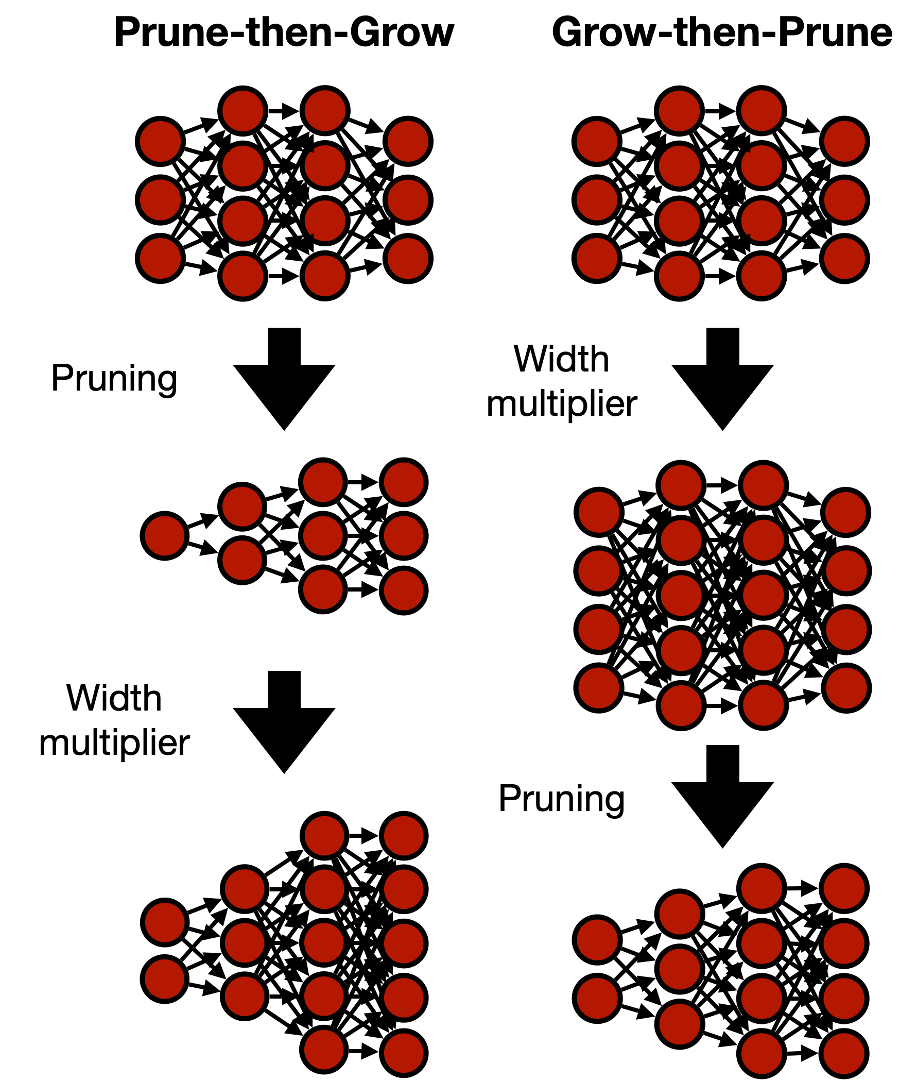}
    \caption{The two width optimization strategies proposed in prior art.}
    \label{fig:wo-methods}
\end{figure}

To optimize the width of a CNN, there are in general two approaches proposed in prior art: \textit{Prune-then-Grow} and \textit{Grow-then-Prune}. \textit{Prune-then-Grow} uses channel pruning methods to arrive at a down-sized CNN with non-trivial layer-wise channel counts followed by re-growing the CNN to its original FLOPs using the width multiplier method~\cite{gordon2018morphnet}. On the other hand, \textit{Grow-then-Prune} first uses the width multiplier method to enlarge the CNN followed by channel pruning methods to trim down channels to match its pre-grown FLOPs~\cite{yu2019autoslim,guo2020dmcp}. The aim of both of these methods is to improve accuracy of the network while maintaining a given FLOPs count. The schematic view of the two approaches is visualized in Figure~\ref{fig:wo-methods}. 

While there are many papers on channel pruning~\cite{li2016pruning,molchanov2019importance}, they mostly focus their analysis on decreasing the inference-time FLOPs of pre-trained models whereas we focus on improving the classification accuracy of a network by optimizing its width while holding inference-time FLOPs constant. While one can use either the \textit{Prune-then-Grow} or \textit{Grow-then-Prune} strategies to arrive at a CNN of equivalent FLOPs, it is not clear if such strategies generally improve performance over the un-optimized baseline as it is not verified in most channel pruning papers. As a result, in this paper, we focus on analyzing algorithms that have demonstrated the effectiveness over the baseline (uniform) width configurations in either \textit{Prune-then-Grow} or \textit{Grow-then-Prune} settings.

\begin{figure*}[t]
    \centering
    \includegraphics[width=0.8\textwidth]{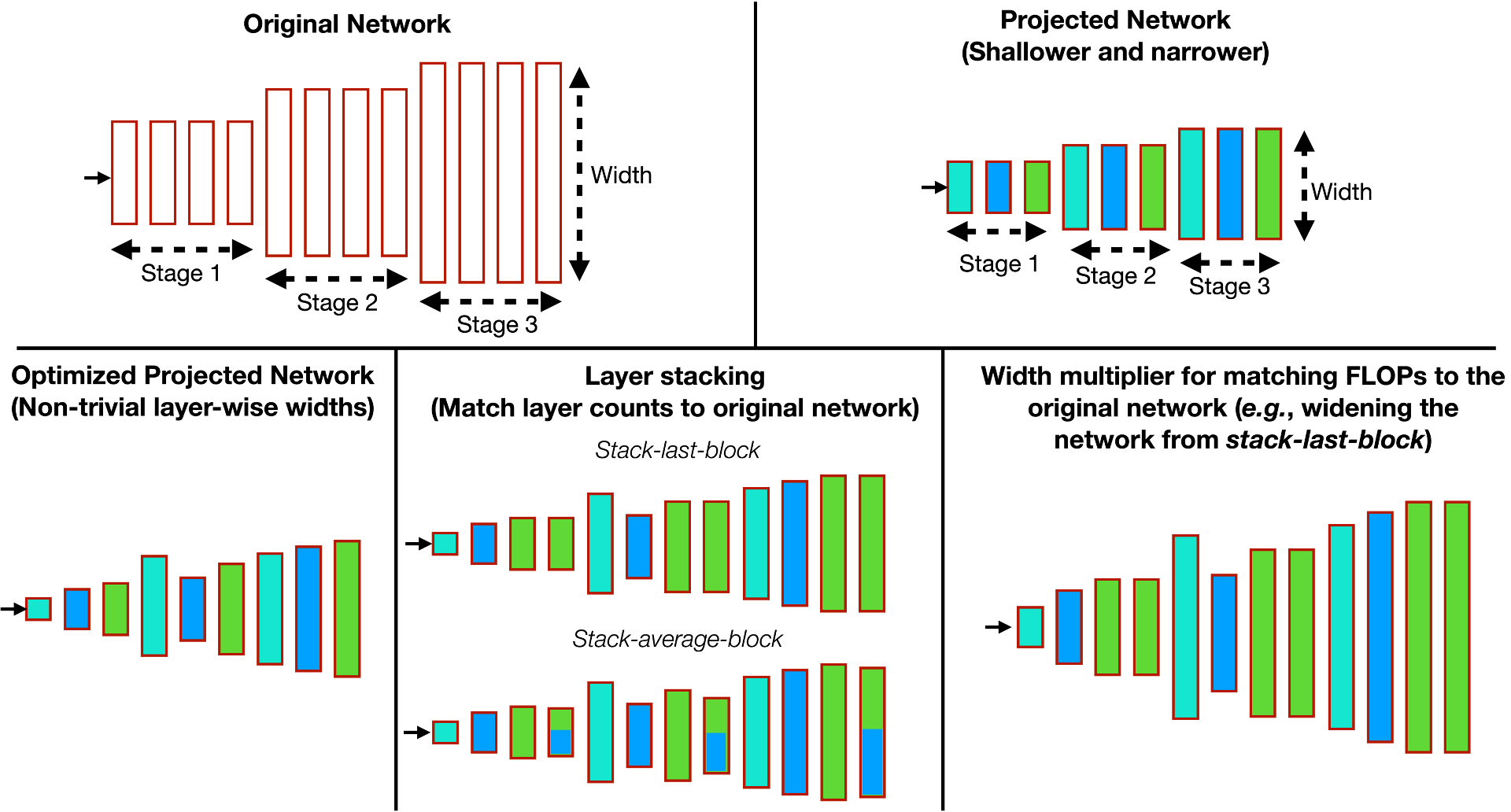}
    \caption{An example for extrapolation. The projected network has fewer layers and channel counts per layer compared to the original network. After width optimization on the projected network, we propose two methods, \textit{i.e.}, stack-last-block and stack-average-block, to match the layer counts to the original network. Finally, we match the FLOPs to the original network with a width multiplier.}
    \label{fig:illustration}
\end{figure*}

\section{Approach}
\subsection{Width optimization methods}
Theoretically, we only care about algorithms $\mathcal{A}$ that ``\textit{solve}'' the width optimization problem (equation~\ref{eq:width_opt}). However, the problem is inherently combinatorially hard. As a result, we use state-of-the-art width optimization algorithms as probes to understand the width optimization problem. More specifically, we consider methods that have reported improved accuracy compared to the un-optimized network given the same test-time FLOPs, and have publicly available code to ensure correctness of implementation. With these criteria, we consider MorphNet~\cite{gordon2018morphnet}, AutoSlim~\cite{yu2019autoslim}, and DMCP~\cite{guo2020dmcp}. Note that we use these algorithms to find the ``network architecture" which will be trained from scratch using the normal training configurations with randomly initialized weights.

\subsection{Projection and extrapolation}\label{sec:extrapolate}
\paragraph{Projection} In projection, there are two aspects: network projection and dataset projection. For network projection, we propose to use the width multiplier to uniformly shrink the channel counts for all the layers to arrive at a narrower model. Additionally, we also propose to use the depth multiplier to uniformly shrink the block counts per each stage of a neural network to arrive at a shallower model. For dataset projection, on the one hand, we propose to sub-sample the training sample in a dataset in a class-balanced way. On the other hand, we propose to sub-sample the spatial dimension of the training images to arrive at images with lower resolutions. When keeping the width optimization algorithm fixed, \textit{i.e.}, training the input network with a certain batch size and training epochs using the input dataset, all the aforementioned projections immediately result in width optimization overhead reduction.

\paragraph{Extrapolation} We consider two aspects for extrapolation: dimension-matching and FLOPs-matching, which are schematically shown in Figure~\ref{fig:illustration}. First, we want the extrapolated network to have the same number of layers as the original network. This is particularly crucial when the original network is projected in the depth dimension, in which case we propose two layer stacking strategies:
\begin{itemize}
    \item \textbf{Stack-last-block}: Stack the width multipliers of the last block of each stage until the desired depth is met. A stage includes convolutional blocks with the same output resolution in the original network. A convolutional block consists of several convolutional layers such as the bottleneck block in ResNet~\cite{he2016deep} and the inverted residual block in MobileNetV2~\cite{sandler2018mobilenetv2}.
    \item \textbf{Stack-average-block}: To avoid mismatches among residual connection, we exclude the first block of each stage and compute the average of the width multipliers across all the rest blocks in a stage, then stack the average width multipliers until the desired depth is met.
\end{itemize}
Note that since existing network designs share the same channel widths for all the blocks in each stage, the above two layer stacking strategies will have the same results when applied to networks with un-optimized widths. 

Second, we want to extrapolate the optimized projected network to a larger one such that it has the FLOPs of the original network. To do so, we propose to use the width multiplier to widen the optimized width. This procedure basically assumes that what determines the optimized widths is the ratio among layers and we show that this assumption is largely correct in Section~\ref{sec:exp} as the optimized widths are largely transferable across network's initial widths.

\section{Experiments}\label{sec:exp}
In this section, we empirically investigate the transferability of the optimized widths across different projection and extrapolation strategies. Specifically, we study projection across architectures by evaluating different widths and depths as well as across dataset properties by dataset sub-sampling and resolution sub-sampling. In addition to analyzing each of these four settings independently, we also investigate a compound projection that involves all four jointly. To measure the transferability, we plot the ImageNet top-1 accuracy of network obtained by direct optimization and width transfer across different projection scales that have different width optimization overhead. Width optimization overhead refers to the FLOPs needed to carry out width optimization. If transferable, we should observe a horizontal line across different width optimization overheads, suggesting that performance is not compromised by deriving the optimized widths from a smaller FLOP configuration. Moreover, we also plot the ImageNet top-1 accuracy for the un-optimized baseline to characterize whether width optimization or width transfer is even useful for some configurations.

\subsection{Experimental Setup}
We used the ImageNet dataset~\cite{deng2009imagenet} throughout the experiments. Unless stated otherwise, we use 224 input resolution. For CNNs, we considered the meta-architecture of ResNet18~\cite{he2016deep} and MobileNetV2~\cite{sandler2018mobilenetv2}. Note that we adjusted the depth and width of ResNet-18 and MobileNetV2 to arrive at a wide variety of models for our width transfer study. Models were each trained on a single machine with 8 V100 GPUs for all the experiments. The width multiplier method applies to all the layers in the CNNs while the depth-multiplier excludes the first and the last stage of MobileNetV2 as there is only one block for each of them. After we obtained the optimized architecture, we trained the corresponding network from scratch with random initialization using the same hyperparameters to analyze their performance. We repeated each experiment three times with different random seeds and reported the mean and standard deviation. Other hyperparameters are detailed in Appendix~\ref{app:hyperparam}.

\subsection{Projection: width}\label{sec:width}
Here, we focus on answering the following question: ``\textit{Do optimized widths obey the channel magnitude invariance?}'' The answer to this question is unclear from existing literature as the current practice is to re-run the optimization across different networks~\cite{guo2020dmcp,gordon2018morphnet,liu2019metapruning}. If the optimized widths are similar across different initial widths, this suggests that the quality of the vector of channel counts are scale-invariant given the current practice of training deep CNNs and the dataset. Additionally, it also has practical benefits where one can use width transfer to reduce the overhead incurred in width optimization. On the other hand, if the optimized widths are dissimilar, this suggests that not only the direction of the vector of channel counts is important, but also its magnitude. That is, for different magnitudes, we may need different orientations. In other words, this suggests that existing practice, though costly, is empirically proved to be necessary.

\begin{figure}[t!]
     \centering
     \begin{subfigure}[b]{0.23\textwidth}
         \centering
         \includegraphics[width=\textwidth]{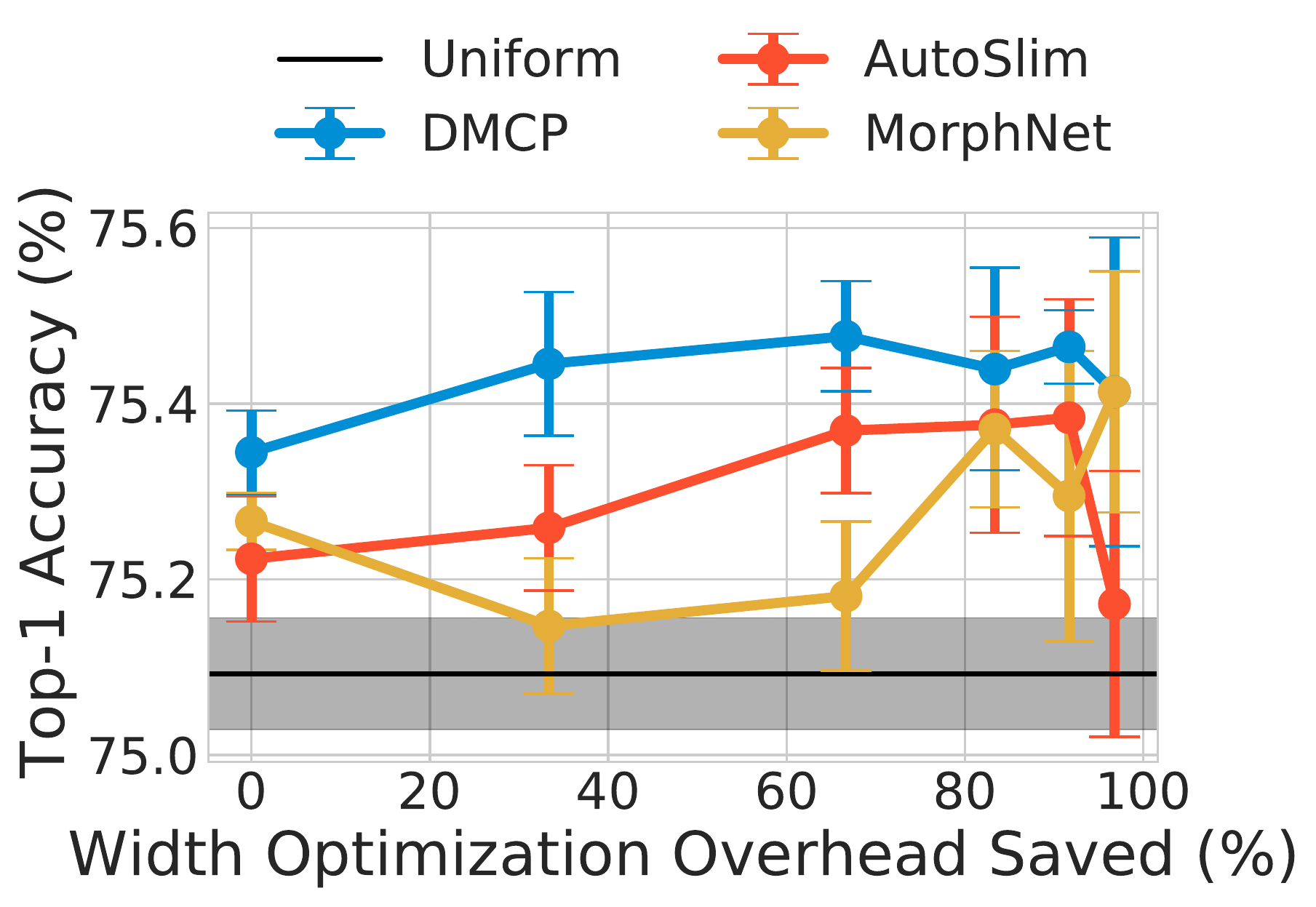}
         \caption{Res18, width}
         \label{width-res18}
     \end{subfigure}
    %  \hfill
     \begin{subfigure}[b]{0.23\textwidth}
         \centering
         \includegraphics[width=\textwidth]{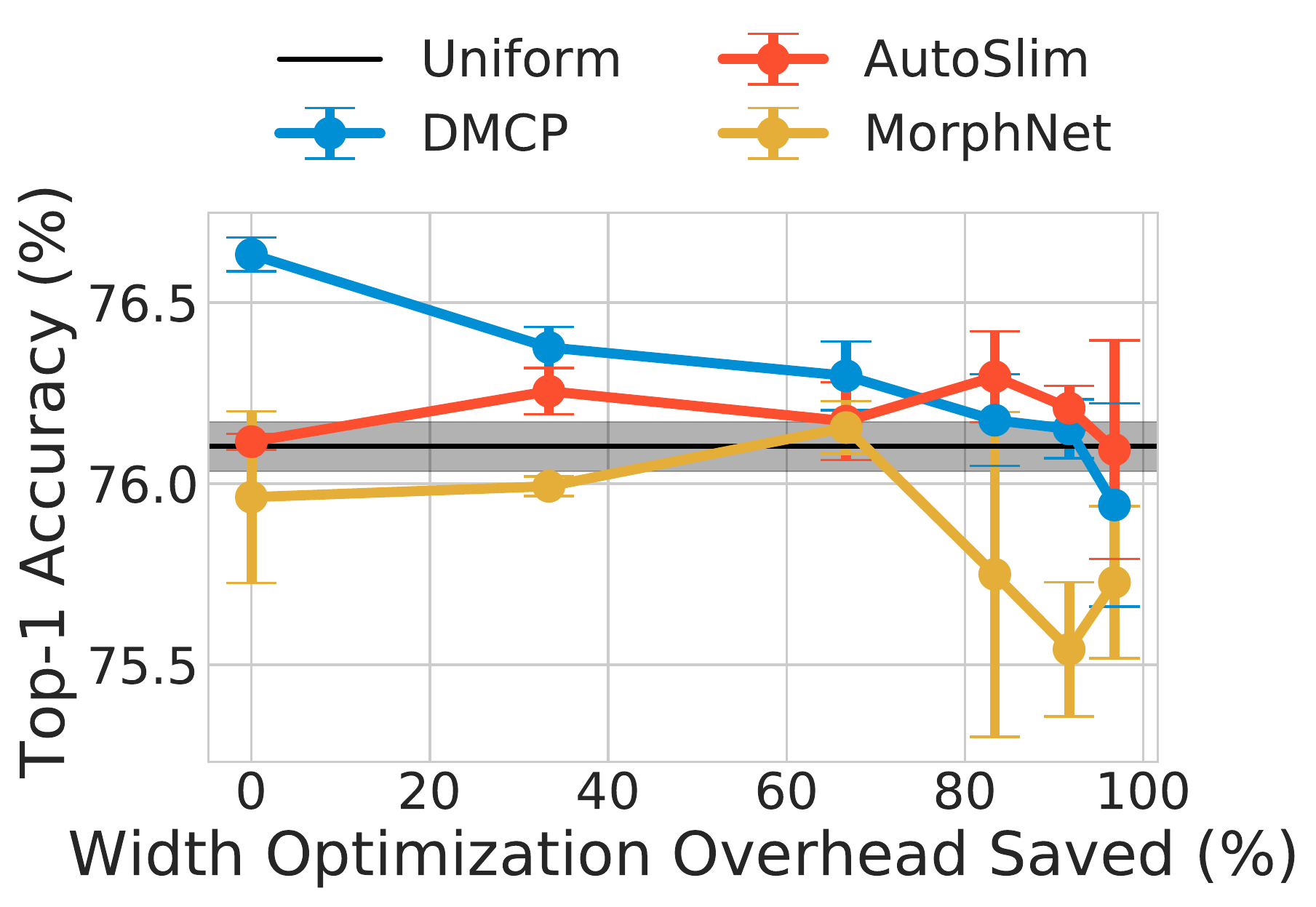}
         \caption{MBv2, width}
         \label{width-mbv2}
     \end{subfigure}
     \begin{subfigure}[b]{0.23\textwidth}
         \centering
         \includegraphics[width=\textwidth]{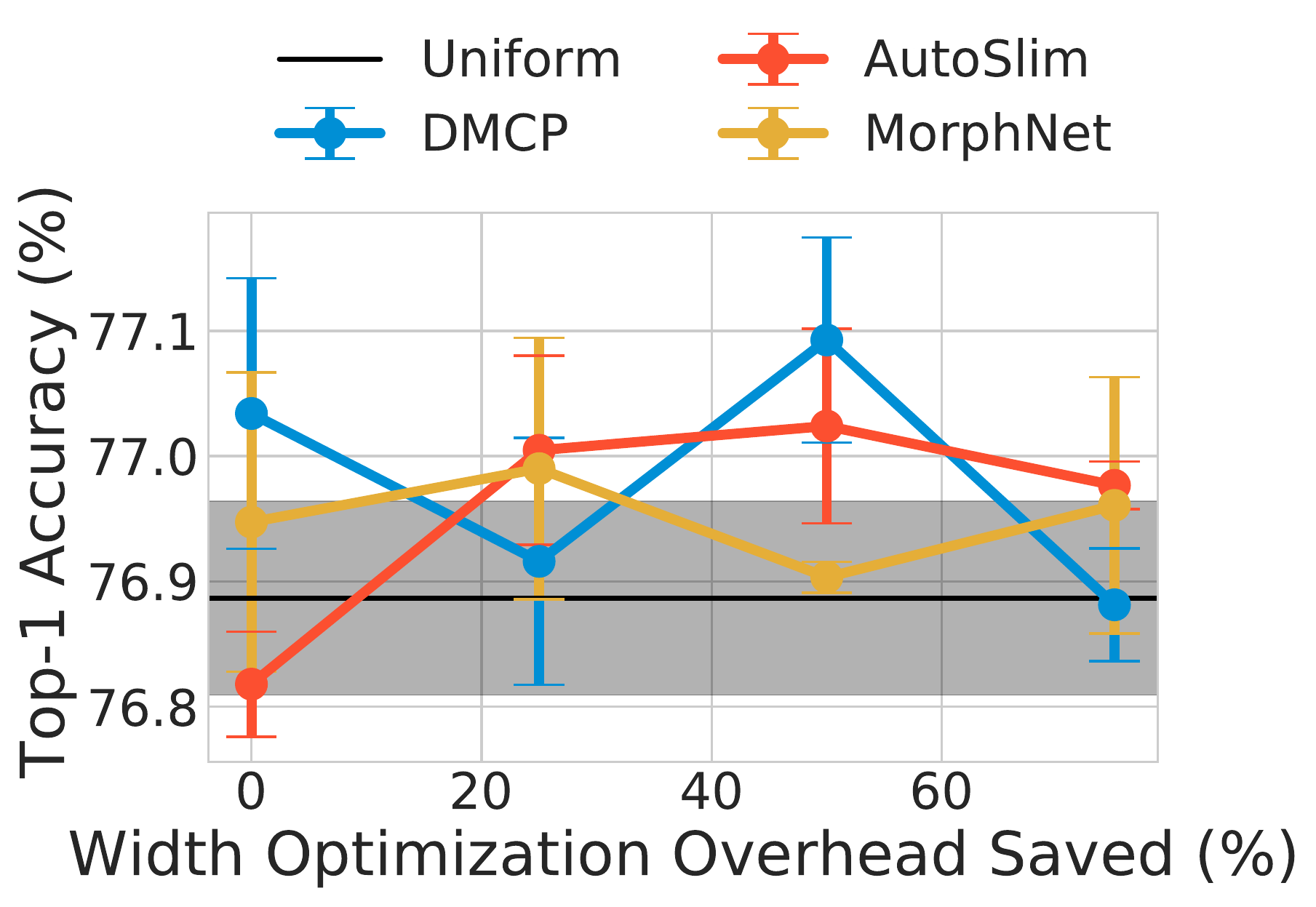}
         \caption{Res18, depth}
         \label{res18-depth-aavg}
     \end{subfigure}
     \hfill
     \begin{subfigure}[b]{0.23\textwidth}
         \centering
         \includegraphics[width=\textwidth]{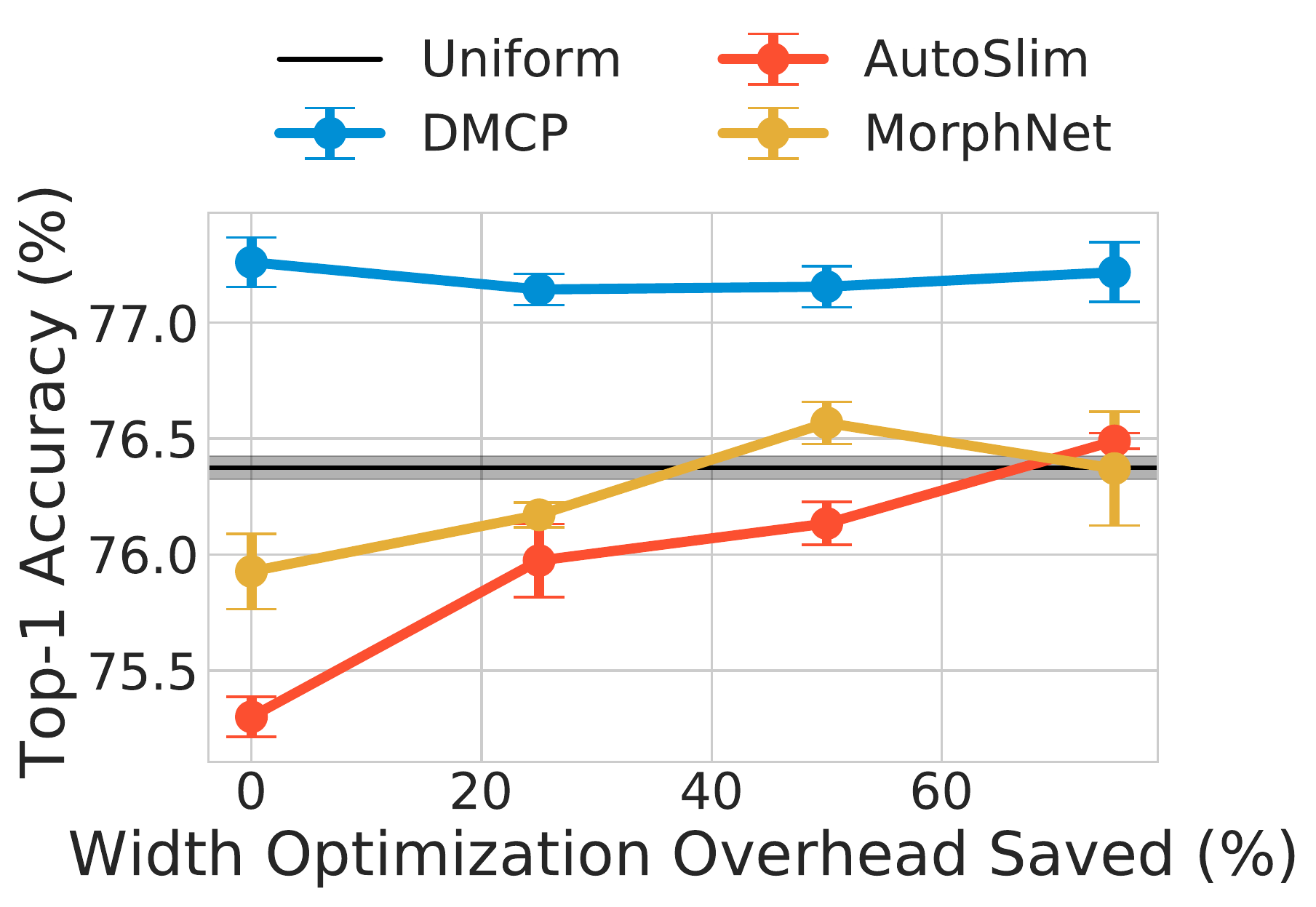}
         \caption{MBv2, depth}
         \label{mbv2-depth-aavg}
     \end{subfigure}
     \caption{Experiments for width transfer under network projection. We plot the ImageNet top-1 accuracy for uniform baseline, width transfer, and direct optimization (the leftmost points). On the x-axis, we plot the width optimization overhead saved by using width transfer.}
    \label{fig:width}
\end{figure}

To empirically study the aforementioned question, we considered the source width multipliers of \{0.312, 0.5, 0.707, 1, 1.414, 1.732\} and the target width multiplier of 1.732, and we analyzed if the source optimized architecture transfers to the target architecture. The set of initial width multipliers is chosen based on square roots of width optimization overhead. We analyzed the transferability in the accuracy space. In Figure~\ref{width-res18} and~\ref{width-mbv2}, we plot the ImageNet top-1 accuracy for the baseline (a $1.732\times$ wide network) and networks obtained by direct optimization and width transfer. For ResNet18, the width optimization overhead can be saved by up to 96\% for all three algorithms without compromising the accuracy improvements gained by the width optimization. 

On MobileNetV2, AutoSlim and MorphNet can transfer well and save up to 80\% width optimization overhead. While DMCP for MobileNetV2 results in 0.4\% top-1 accuracy loss when using width transfer, the transferred width can still outperform the uniform baseline, which is encouraging for applications that allow such accuracy degradation in exchange for 83\% width optimization overhead savings. More specifically, that would reduce compute time from 160 GPU-hours all the way to 30 GPU-hours for MobileNetV2 measured using a batch size of 1024, a major saving. Our results suggest that a good orientation for the optimized channel vector continues to be suitable across a wide range of magnitudes.

\begin{figure}[t]
    \centering
     \begin{subfigure}[b]{0.33\textwidth}
         \centering
         \includegraphics[width=\textwidth]{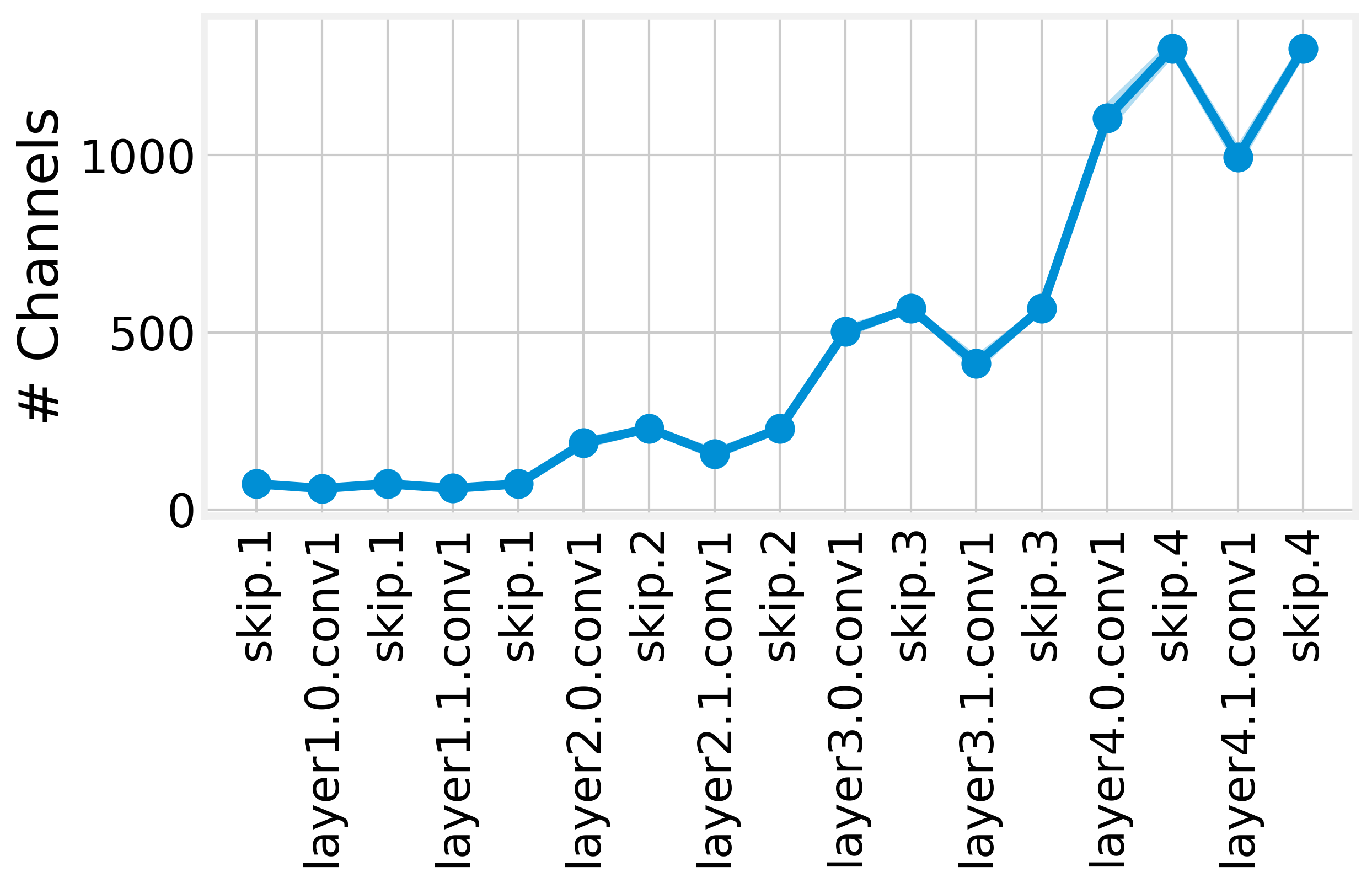}
         \caption{ResNet18}
     \end{subfigure}
     \hfill
     \begin{subfigure}[b]{0.33\textwidth}
         \centering
         \includegraphics[width=\textwidth]{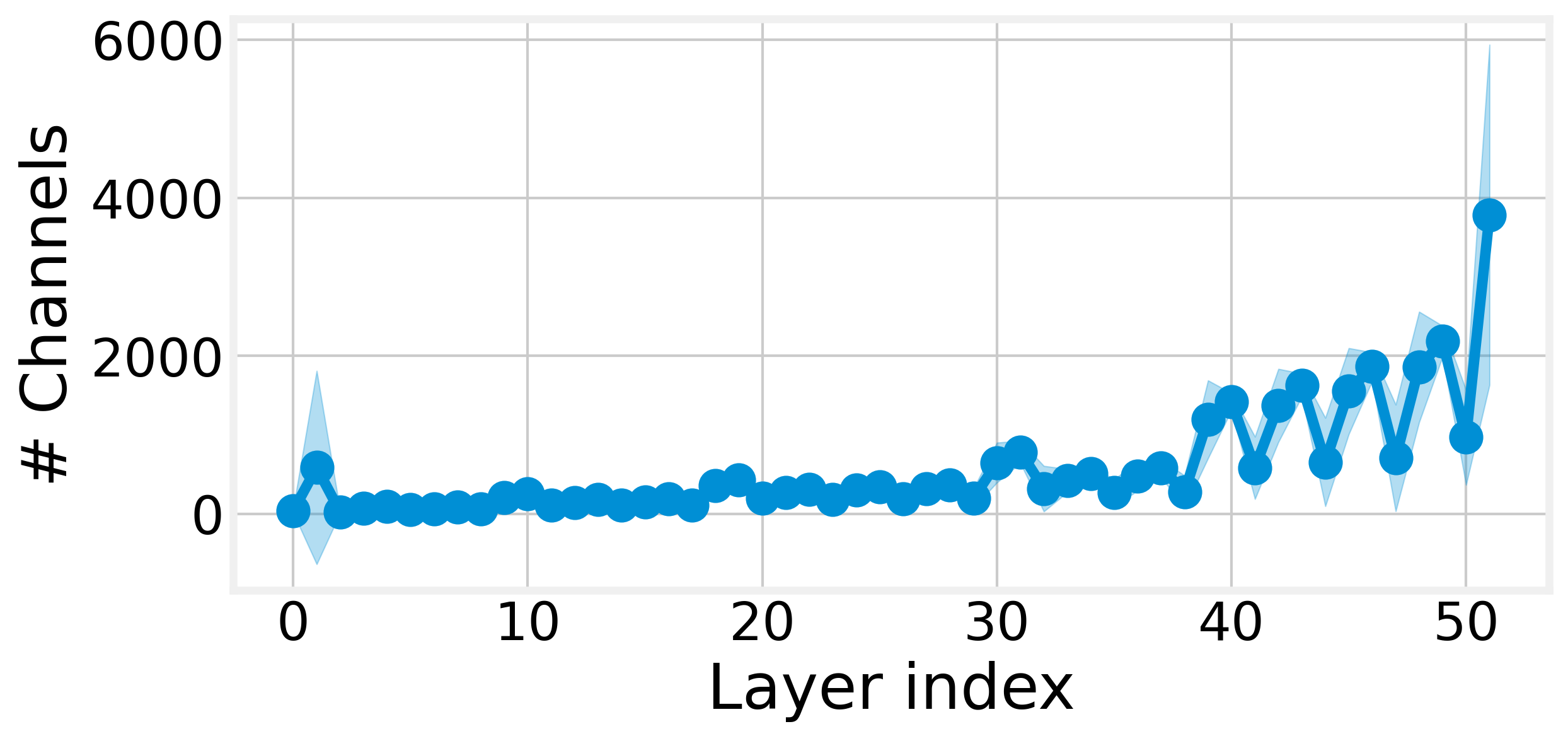}
         \caption{MobileNetV2}
     \end{subfigure}
     \caption{The average optimized width for ResNet18 and MobileNetV2. They are averaged across the optimized widths. We plot the mean in solid line with shaded area representing standard deviation.}
    \label{fig:avg-width}
\end{figure}

Since the optimized widths are highly transferable, we are interested in the resulting widths for both CNNs. We find that the later layers tend to increase a lot compared to the un-optimized ones. Concretely, in un-optimized networks, ResNet18 has 512 channels in the last layer and MobileNetV2 has 1280 channels in the last layer. In contrast, the average optimized width has 1300 channels for ResNet18 and 3785 channels for MobileNetV2. We visualize the average widths for ResNet18 and MobileNetV2 (average across optimized widths) in Figure~\ref{fig:avg-width}.

\subsection{Projection: depth}\label{sec:depth}

\begin{figure}[h]
     \centering
     \begin{subfigure}[b]{0.23\textwidth}
         \centering
         \includegraphics[width=\textwidth]{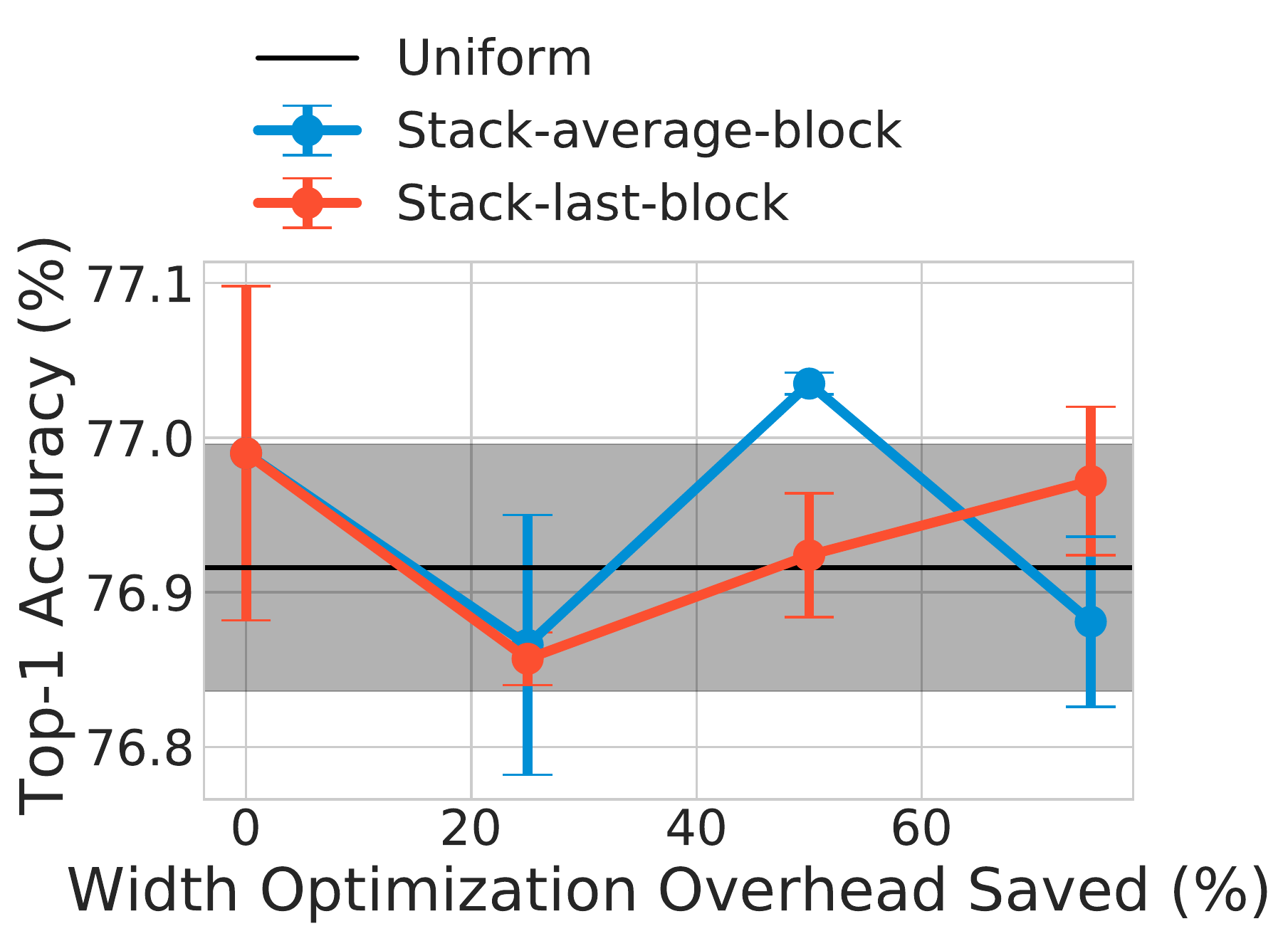}
         \caption{ResNet18}
     \end{subfigure}
     \hfill
     \begin{subfigure}[b]{0.23\textwidth}
         \centering
         \includegraphics[width=\textwidth]{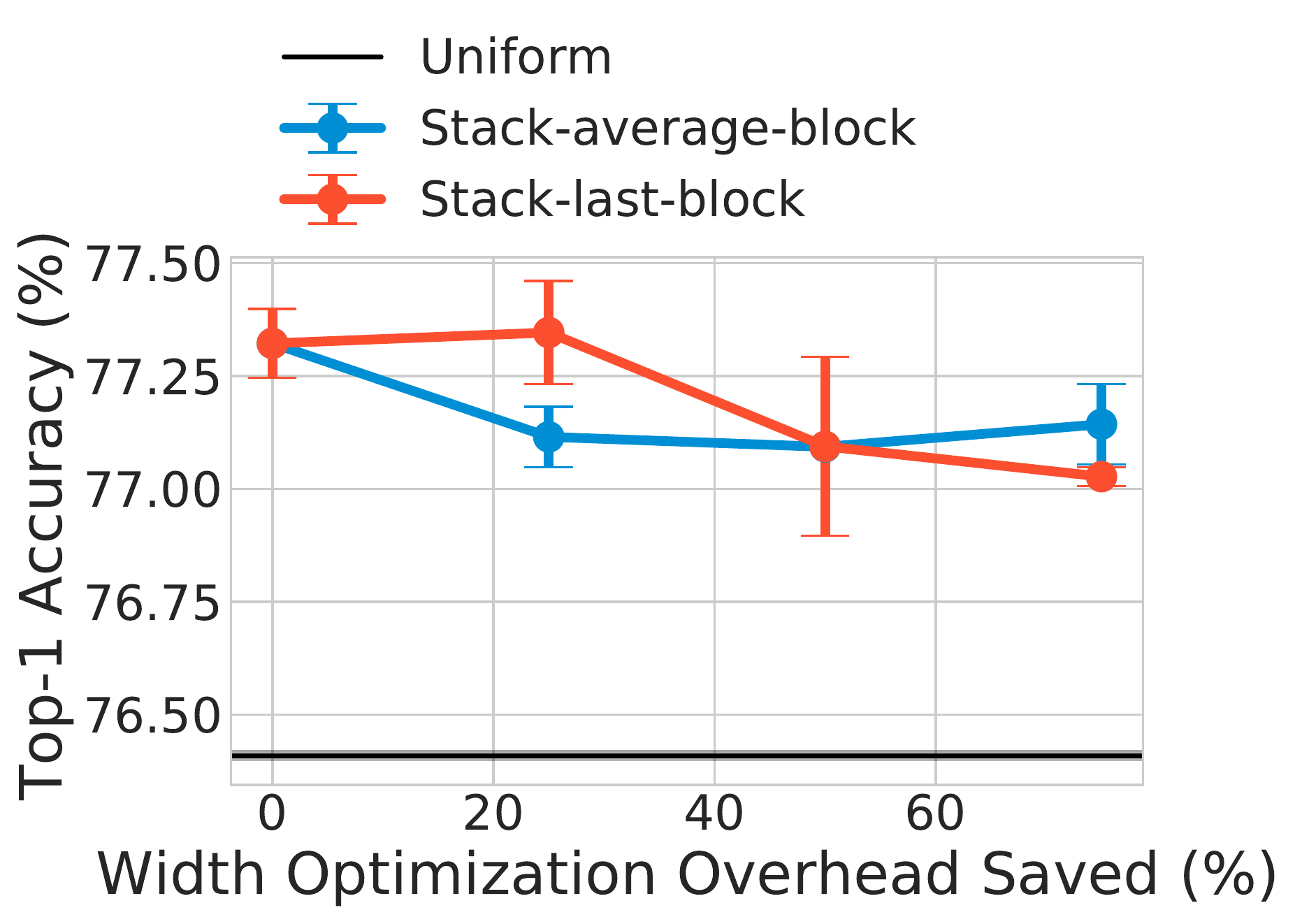}
         \caption{MobileNetV2}
     \end{subfigure}
     \caption{We compare the two layer-stacking strategies using DMCP for both ResNet18 and MobileNetV2. We can observe that both stack-average-block and stack-last-block perform similarly.
    }
    \label{fig:depth-methods}
\end{figure}

Next, we asked whether networks with different initial depths share common structure in their optimized widths. Specifically, ``\textit{Do the optimized widths obey the within-stage channel counts invariance?}" Because making a network deeper will add new layers with no corresponding optimized width, we will need a mechanism to map the vector optimized widths to a vector with far more elements. Here, we first compared the two layer-stacking methods proposed in Section~\ref{sec:extrapolate} using DMCP for ResNet18 and MobileNetV2. As shown in Figure~\ref{fig:depth-methods}, both \textit{stack-last-block} and \textit{stack-average-block} layer stacking strategies perform similarly. As a result, we use \textit{stack-average-block} for all other experiments. We considered \{1, 2, 3, 4\} as the source depth multipliers and use 4 as the target depth multiplier. Similar to the analysis done in Section~\ref{sec:width}, we analyzed the similarity in the accuracy space.

As shown in Figure~\ref{res18-depth-aavg} and~\ref{mbv2-depth-aavg}, we find that the optimized widths stay competitive via simple layer stacking methods and up to 75\% width optimization overhead can be saved if we were to optimize the width using width transfer for all three algorithms and two networks. This finding also suggests that the relative values of optimized widths share common structure across networks that differ in depth. In other words, the pattern of width multipliers across depth is scale-invariant. Interestingly, we find that width transfer \textit{improves} direct optimization in terms of accuracy when it comes to AutoSlim and MorphNet as we see a positive slope for these two methods on both networks. We conjecture that this is due to both AutoSlim and MorphNet are affected more by the dimensionality\footnote{Width depth projection, we effectively reduce the dimensionality of the search problem.} of the problem (the number of widths to be learned), and that the within-stage channel invariance largely holds.

\subsection{Projection: resolution}
The input resolution and the channel counts of a CNN are known to be related when it comes to the test accuracy of a CNN. As an example, it is known empirically that a wider CNN can benefit from inputs with a higher resolution than a narrower net can~\cite{tan2019efficientnet}. As a result, it is not clear \textit{if the optimized widths obey the spatial resolution invariance}. If the optimized widths indeed obey the spatial resolution invariance, this suggests that although wider networks benefit more from a higher resolution inputs, the non-uniform widths that result in better performance are similar. On the other hand, if the optimized widths are different, it suggests that, when it comes to the test accuracy, the relationship between channel counts and input resolution is more involved than the level of over-parameterization.

\begin{figure}[t!]
     \centering
     \begin{subfigure}[b]{0.23\textwidth}
         \centering
         \includegraphics[width=\textwidth]{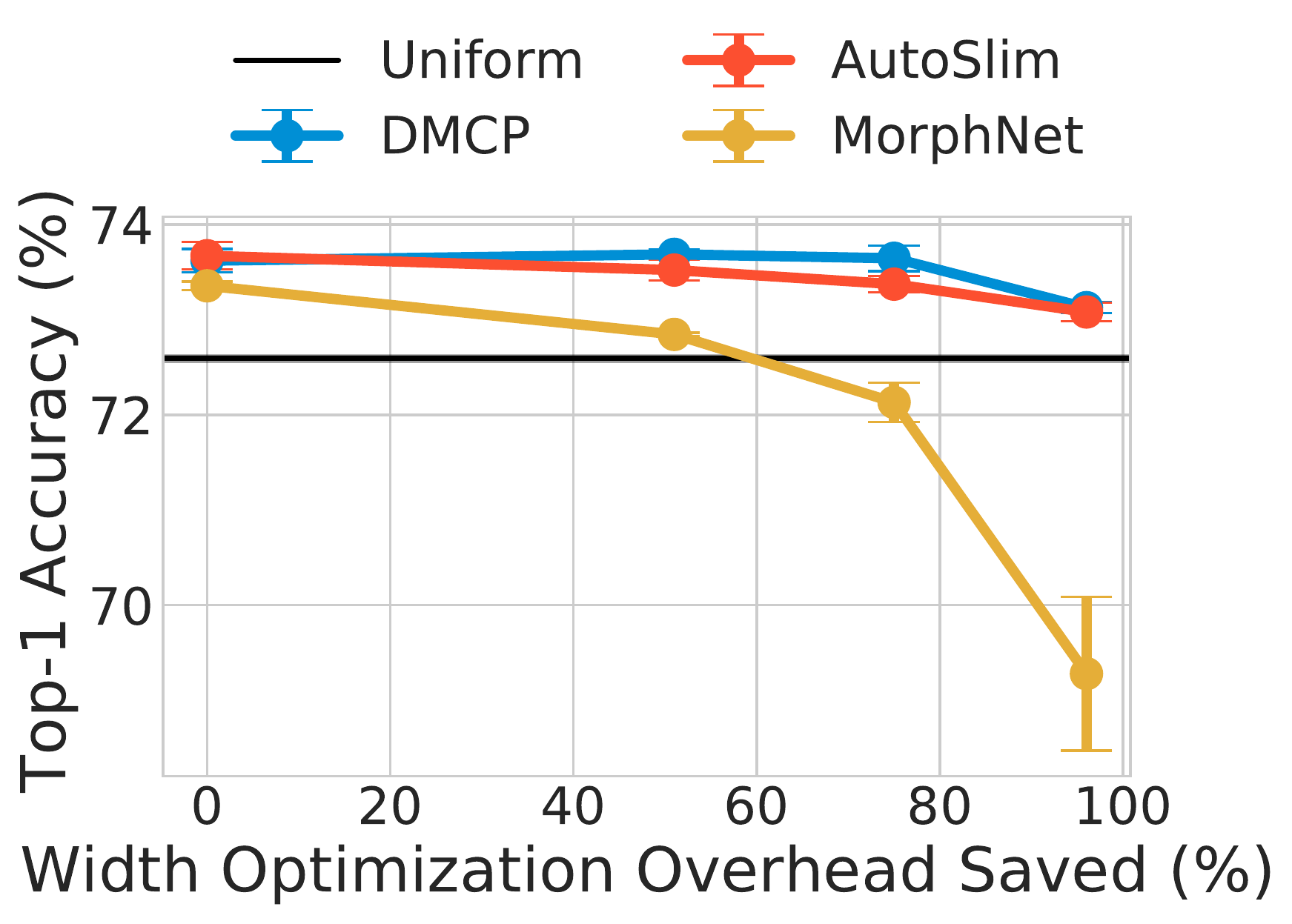}
         \caption{Res18, resolution}
         \label{res-res18}
     \end{subfigure}
    %  \hfill
     \begin{subfigure}[b]{0.23\textwidth}
         \centering
         \includegraphics[width=\textwidth]{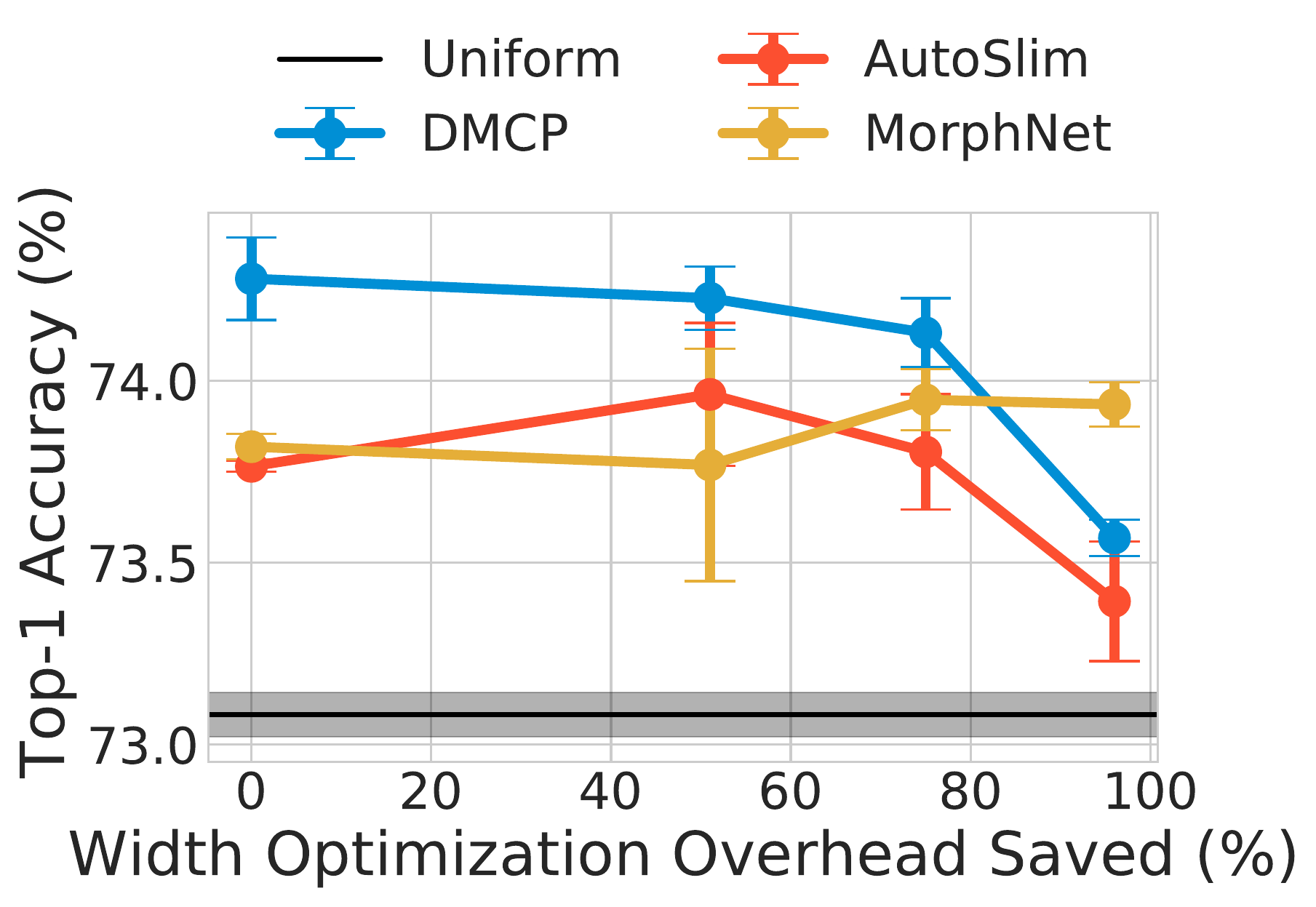}
         \caption{MBv2, resolution}
         \label{res-mbv2}
     \end{subfigure}
     \begin{subfigure}[b]{0.23\textwidth}
         \centering
         \includegraphics[width=\textwidth]{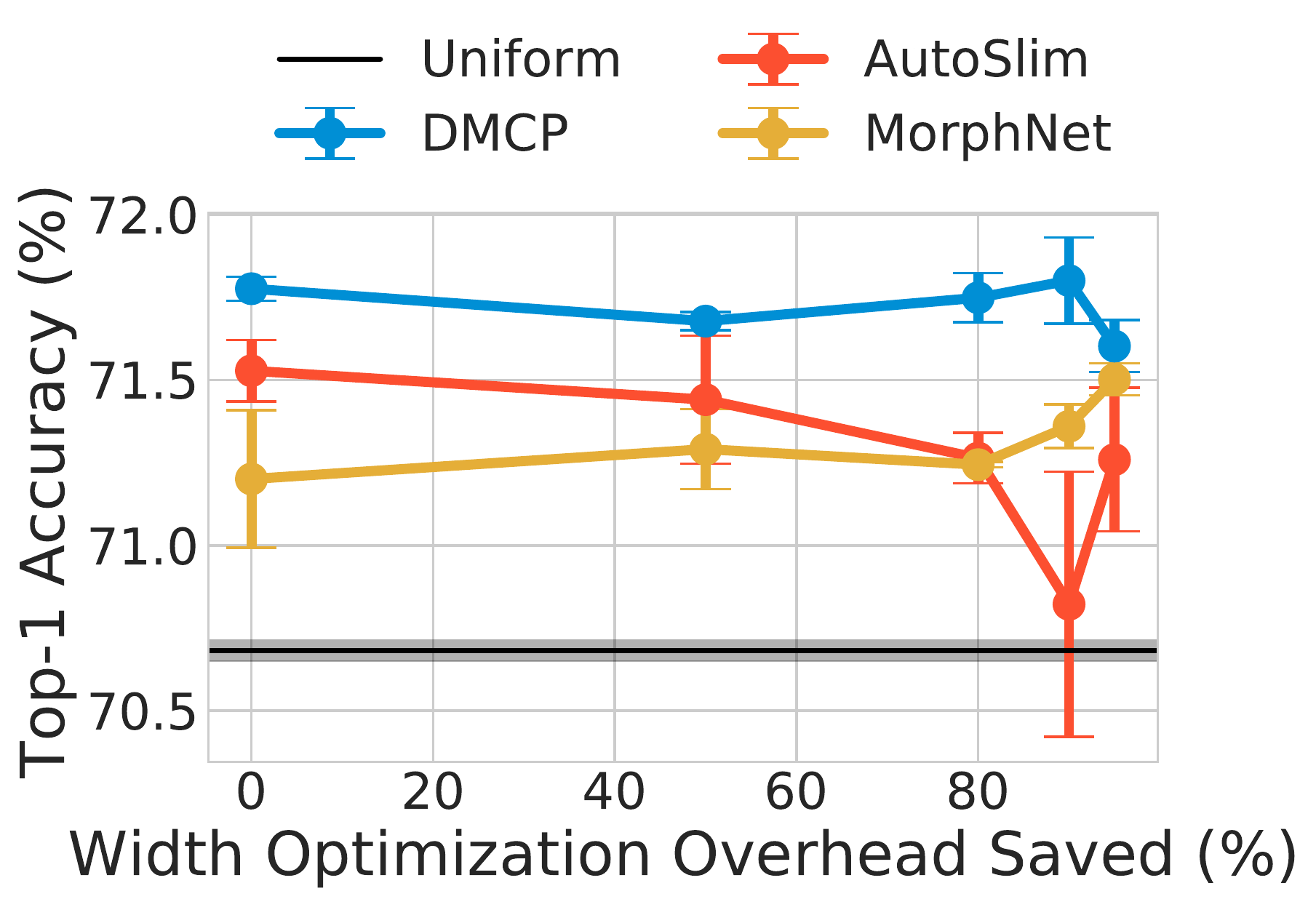}
         \caption{Res18, dataset size}
         \label{ds-res18}
     \end{subfigure}
    %  \hfill
     \begin{subfigure}[b]{0.23\textwidth}
         \centering
         \includegraphics[width=\textwidth]{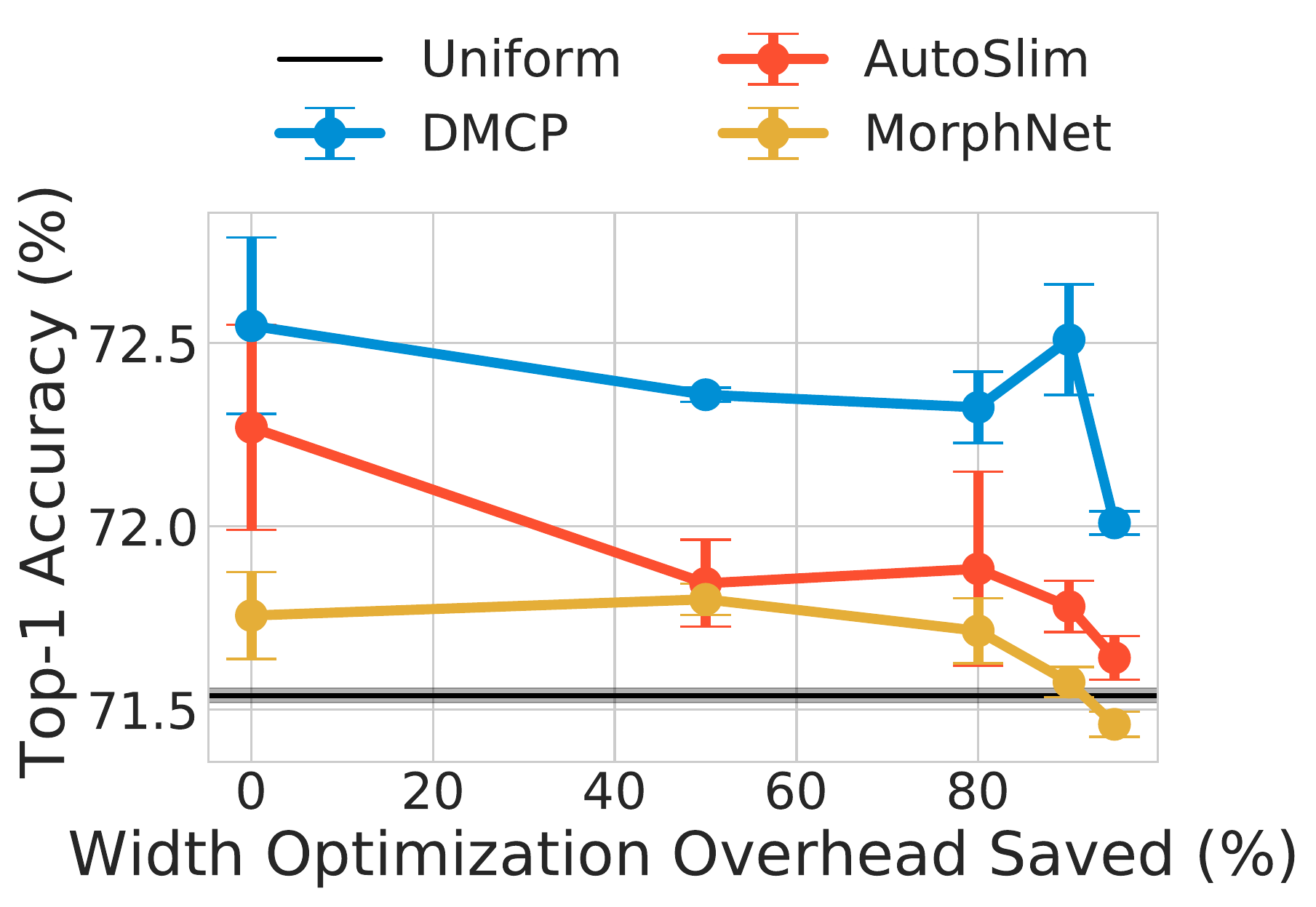}
         \caption{MBv2, dataset size}
         \label{ds-mbv2}
     \end{subfigure}
    \caption{Experiments for width transfer under dataset projection. We plot the ImageNet top-1 accuracy for uniform baseline, width transfer, and direct optimization (the leftmost points). On the x-axis, we plot the width optimization overhead saved by using width transfer.}
    \label{fig:dataset}
\end{figure}

To study the aforementioned question, we considered the input resolution for the source to be \{64, 160, 224, 320\} and choose a target of 320. As shown in Figure~\ref{res-res18} and~\ref{res-mbv2}, we find that except for MorphNet targeting ResNet18, all other algorithm and network combinations can achieve up to 96\% width optimization overhead savings with the optimized widths that are still better than the uniform baseline. By saving 75\% width optimization overhead, we can stay close to the performance obtained via direct optimization. Interestingly, we find that MorphNet had a very different optimized widths when transferred from resolution 64 for ResNet18, which leads to the worse performance for ResNet18 compared to direct optimization. The similarity among the optimized widths are detailed in Figure~\ref{fig:sim} in Appendix.

\subsection{Projection: dataset size}\label{sec:ds}
The dataset size is often critical for understanding the generalization performance of a learning algorithm. Here, we would like to understand \textit{if the optimized widths obey the sample size invariance}. We considered sub-sampling the ImageNet dataset to result in a source of $\{5\%, 10\%, 20\%, 50\%, 100\%\}$ of the original training data. Similar to previous analysis, we tried to transfer the optimized widths obtained using the smaller configurations to the largest configuration, \textit{i.e.}, 100\% of the original training data. As shown in Figure~\ref{ds-res18} and~\ref{ds-mbv2}, widths optimized for smaller dataset sizes transfer well to large dataset sizes. That is, 95\% width optimization overhead can be saved and still outperforms the uniform baseline for both networks. On the other hand, 90\% width optimization overhead can be saved and can still match the performance of direct optimization for DMCP. This suggests that the amount of training data barely affects width optimization, especially for DMCP, which is surprising.

\begin{table*}[t]
    \centering
    \resizebox{0.8\textwidth}{!}{%
        \begin{tabular}{c|c|c|c|c}
            Network & Baseline & DMCP & Width transfer & Overhead (direct$\rightarrow$width transfer)\\
            \hline
            ResNet50 & $77.97\pm0.09$ & $78.23\pm0.11$ & $78.07\pm0.12$ & $37.4\rightarrow 1.3$\\
            ResNet101 & $79.43\pm0.07$ & $79.70\pm0.05$ & $79.54\pm0.07$ & $66.7\rightarrow 2.7$\\
            EfficientNetB3 & $80.02\pm0.09$ & $80.24\pm0.02$ & $80.19\pm0.10$ & $80\rightarrow 3$
        \end{tabular}
    }
    \caption{Compound width transfer for other CNNs. Width optimization overhead measured with 8 NVIDIA V100 GPUS on a single machine.}
    \label{tab:large}
\end{table*}

\subsection{Compound projection}
\begin{figure}[h]
     \centering
     \begin{subfigure}[b]{0.23\textwidth}
         \centering
         \includegraphics[width=\textwidth]{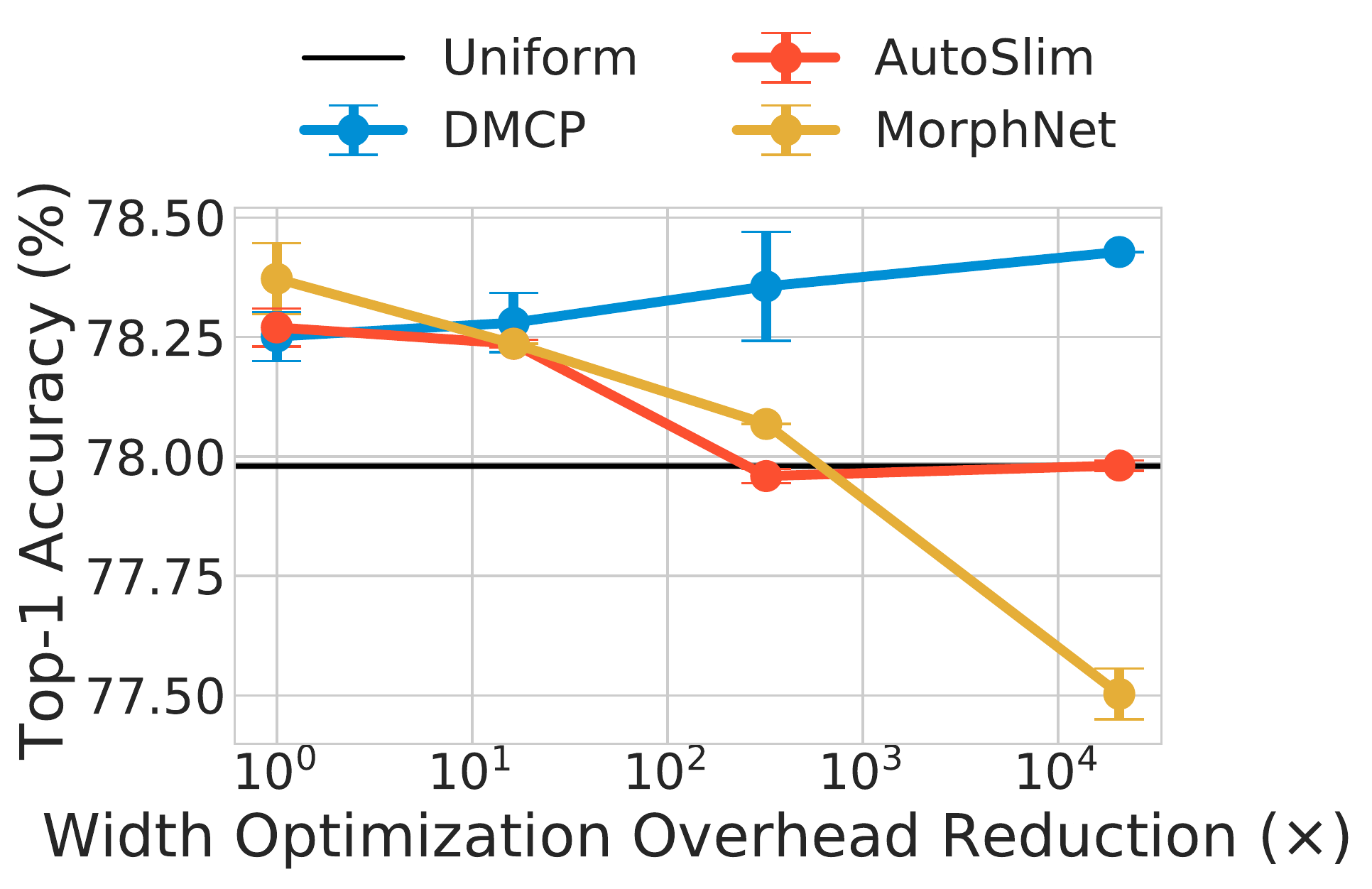}
         \caption{ResNet18}
         \label{compound-res18}
     \end{subfigure}
     \hfill
     \begin{subfigure}[b]{0.23\textwidth}
         \centering
         \includegraphics[width=\textwidth]{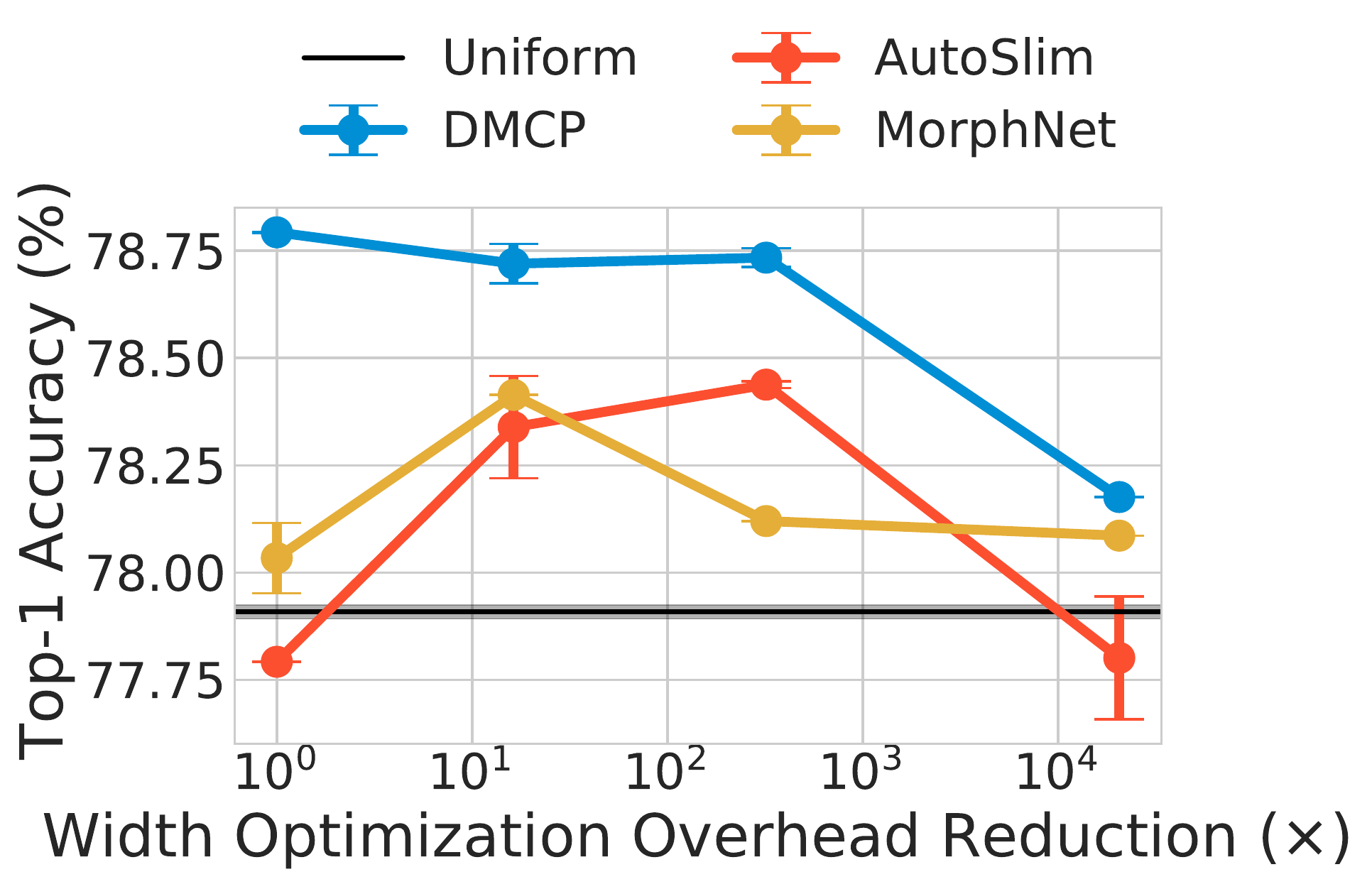}
         \caption{MobileNetV2}
         \label{compound-mbv2}
     \end{subfigure}
    \caption{Width transfer with compound projection.}
    \label{fig:compound}
\end{figure}
From previous analyses, we find that the optimized widths are largely transferable across various projection methods independently. Here, we further empirically analyzed if the optimized width can be transferable across compound projection. To do so, we considered linearly interpolating all four projection methods and analyzed if the width optimized using cost-efficient settings can transfer to the most costly setting. Specifically, let a tuple (width, depth, resolution, dataset size) denote a training configuration. We considered the source to be \{(0.312,1,64,5\%), (0.707,1,160,10\%), (1,1,224,50\%), (1.414,2,320,100\%)\} and the target to be (1.414,2,320,100\%). As shown in Figure~\ref{fig:compound}, the optimized width is transferable across compound projection. Specifically, we can achieve up to $320\times$ width optimization overhead reduction with width transfer for the best performing algorithm, DMCP. Additionally, it also suggests that the four projection dimensions are not tightly coupled for width optimization.

\paragraph{Applications to other target CNNs} We considered using compound width transfer for ResNet50, ResNet101, and EfficientNetB3. For projection, we consider the width, depth, resolution and dataset size to be 0.707, 0.5, 160, and 20\%, respectively. As shown in Table~\ref{tab:large}, up to $30\times$ wall-clock time reduction is achieved with less than $0.2\%$ top-1 accuracy degradation. Considering a scenario where one wants to optimize the width of a network and train such a network for deployment. Width optimization reduce the overall training cost from $3\times$ to $1.06\times$. Such a huge optimization cost reduction can enable fast exploration for the benefits of width optimization for large models without paying the considerable costs.

\subsection{Comparing to cheap pruning methods}
\begin{figure}[t!]
    \centering
    \includegraphics[width=0.3\textwidth]{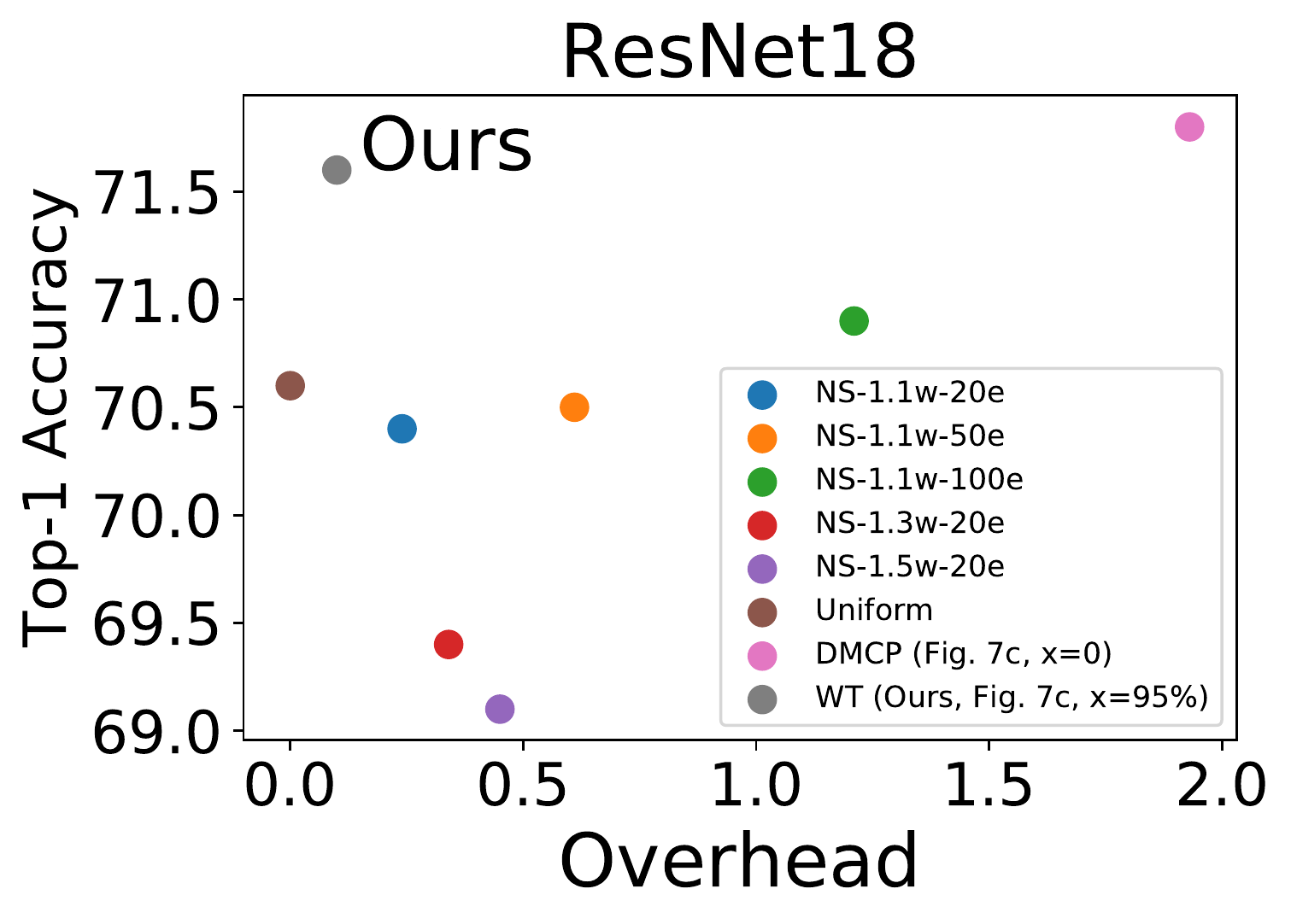}
    \caption{Comparing the proposed using DMCP with width transfer, DMCP, and network slimming.} %\ari{Need to make text in figure legend bigger (or ideally, reduce most of the unnecessary whitespace around figures and make whole figure bigger). Very hard to read without really zooming in. Also, swap a and b since b comes first in text. Also maybe add text in grey saying "Ours" next to the gray dot to make it obvious that's ours. For b, not clear which is source and which is target.}}
    \label{fig:reg_lt}
\end{figure}

While adopting state-of-the-art channel optimization directly can be costly, one may consider using cheap pruning methods and adopt a Prune-then-Grow strategy to carry out width optimization. We compare to magnitude-based channel pruning: network slimming (NS)~\cite{liu2017learning} that prunes channels based on the magnitude of $\gamma$ of the batch normalization layer. NS-$x$w-$y$e follows a three-step procedure: train an $x\times$ wider network for $y$ epochs, prune the network with global $\gamma$ ranking, and re-train the pruned network using full training schedule. The induced overhead for width optimization lies in the first step. The comparisons with NS for ResNet18 on ImageNet is shown in Fig.~\ref{fig:reg_lt}. Using DMCP directly is indeed the most expensive one, but it has the best performance. Our width transfer achieves similar performance compared to DMCP with overhead lower than magnitude-based pruning.

\section{Related work}
\paragraph{Neural architecture search}
Our work is tightly connected to understanding the transferability of the searched results from neural architecture search (NAS) algorithms where the search space is determined by the layer-wise channel counts of a seed network. The transferability of NAS has been recently explored in several papers considering different search spaces and perspectives. Zoph \textit{et al.}\cite{Zoph_2018_CVPR} have proposed to search the best cell on a small dataset and use the searched cell on a large dataset. Panda \textit{et al.}\cite{panda2020nastransfer} have analyzed the transferability of the solutions of various NAS algorithms in the DARTS search space~\cite{liu2018darts} and have concluded that the design of the proxy datasets for the search has a great impact on the transferability of the searched result. Critically however, prior work has neglected to include the channel width multipliers in the search space, instead only focusing on proxy datasets~\cite{panda2020nastransfer,Zoph_2018_CVPR}. Consequentially, the relationship between optimized widths across different architectures has not been examined previously. Others~\cite{wong2018transfer,casale2019probabilistic,lu2020nsganetv2} have analyzed the transferability of the search processes as opposed to the searched solutions. The key difference among the two is that the transfer of the search processes is algorithm-dependent while the transfer of the searched solutions is not. Given width optimization is a fast-developing field, we study the transferability in the solution space to have a more general result.

EcoNAS~\cite{zhou2020econas} is closely related to our work as it also systematically investigated several proxy training configurations for neural architecture search. However, the crucial difference between EcoNAS and ours is that, in our study, the number of channels is not only a projection dimension, but also the optimization variable. As a result, the extrapolation step is necessary for our study but not for EcoNAS. Consequentially, the proposed width transfer has a dimensionality reduction effect for hyperparameter optimization (as noted in Section 4.3) while EcoNAS does not. Lastly, the search space for EcoNAS is the cell-based search space while our channel search space is orthogonal to theirs and can be applied to any network for architectural fine-tuning.

\paragraph{Channel pruning}
Channel pruning is an active research topic for efficient network design. More specifically, channel pruning determines how we can prune an existing network in the channel dimension so as to retain the most accuracy~\cite{chin2020towards,he2018soft,he2018amc,li2016pruning,molchanov2019importance}. A channel pruning procedure often has the weights and the optimized channel counts coupled together. Inspired by Liu \textit{et al.}~\cite{liu2018rethinking}, who had empirically shown the importance of the optimized channel counts, we take a step further by understanding the transferability of the searched channel counts across different input network and dataset transformations.

\section{Conclusion}
In this paper, we take a first step in understanding the transferability of the optimized widths across different width optimization algorithms and invariance dimensions. Our empirical analysis sheds light on the structure of the width optimization problem, which can be used to design better optimization methods. More specifically, by exploiting the channel magnitude and within-stage channel counts invariances, we not only can reduce the computational cost needed to width optimization, but also reduce the dimension of the optimization variables\footnote{In the channel search space, we have one optimization variable for each layer.}. Per our analysis, we can achieve up to $320\times$ reduction in width optimization overhead without compromising the top-1 accuracy on ImageNet.

\section*{Acknowledgement}
This research was supported in part by NSF CCF Grant No. 1815899, and NSF CSR Grant No. 1815780.

% In this paper, we take a first step in understanding the transferability of the optimized widths across different width optimization algorithms and projection dimensions. This investigation sheds light on the width optimization problem, which is often regarded as a black-box problem and tackled with general optimization methods~\cite{liu2019metapruning,guo2020dmcp,gordon2018morphnet}. More specifically, we show that there are common structures in the optimized widths obtained across a wide range of settings such that one can successfully transfer the optimized width to various settings with competitive performance. Per our analysis, we can achieve up to $320\times$ reduction in width optimization overhead without compromising the top-1 accuracy on ImageNet. Our findings not only suggest an efficient alternative to conduct width optimization, but also imply that width optimization can be done for lower dimensional inputs, which can be beneficial since it allows a more effective traversal of the design space.

{\small
\bibliographystyle{ieee_fullname}
\bibliography{egbib}

\begin{thebibliography}{10}\itemsep=-1pt

\bibitem{casale2019probabilistic}
Francesco~Paolo Casale, Jonathan Gordon, and Nicolo Fusi.
\newblock Probabilistic neural architecture search.
\newblock {\em arXiv preprint arXiv:1902.05116}, 2019.

\bibitem{chin2020one}
Ting-Wu Chin, Pierce I-Jen Chuang, Vikas Chandra, and Diana Marculescu.
\newblock One weight bitwidth to rule them all.
\newblock In Adrien Bartoli and Andrea Fusiello, editors, {\em Computer Vision
  -- ECCV 2020 Workshops}, pages 85--103, Cham, 2020. Springer International
  Publishing.

\bibitem{chin2020towards}
Ting-Wu Chin, Ruizhou Ding, Cha Zhang, and Diana Marculescu.
\newblock Towards efficient model compression via learned global ranking.
\newblock In {\em Proceedings of the IEEE/CVF Conference on Computer Vision and
  Pattern Recognition}, pages 1518--1528, 2020.

\bibitem{chin2020pareco}
Ting-Wu Chin, Ari~S Morcos, and Diana Marculescu.
\newblock Pareco: Pareto-aware channel optimization for slimmable neural
  networks.
\newblock {\em arXiv preprint arXiv:2007.11752}, 2020.

\bibitem{cubuk2020randaugment}
Ekin~D Cubuk, Barret Zoph, Jonathon Shlens, and Quoc~V Le.
\newblock Randaugment: Practical automated data augmentation with a reduced
  search space.
\newblock In {\em Proceedings of the IEEE/CVF Conference on Computer Vision and
  Pattern Recognition Workshops}, pages 702--703, 2020.

\bibitem{deng2009imagenet}
Jia Deng, Wei Dong, Richard Socher, Li-Jia Li, Kai Li, and Li Fei-Fei.
\newblock Imagenet: A large-scale hierarchical image database.
\newblock In {\em 2009 IEEE conference on computer vision and pattern
  recognition}, pages 248--255. Ieee, 2009.

\bibitem{ding2019regularizing}
Ruizhou Ding, Ting-Wu Chin, Zeye Liu, and Diana Marculescu.
\newblock Regularizing activation distribution for training binarized deep
  networks.
\newblock In {\em Proceedings of the IEEE Conference on Computer Vision and
  Pattern Recognition}, pages 11408--11417, 2019.

\bibitem{dong2019network}
Xuanyi Dong and Yi Yang.
\newblock Network pruning via transformable architecture search.
\newblock In {\em Advances in Neural Information Processing Systems}, pages
  760--771, 2019.

\bibitem{gordon2018morphnet}
Ariel Gordon, Elad Eban, Ofir Nachum, Bo Chen, Hao Wu, Tien-Ju Yang, and Edward
  Choi.
\newblock Morphnet: Fast \& simple resource-constrained structure learning of
  deep networks.
\newblock In {\em Proceedings of the IEEE conference on computer vision and
  pattern recognition}, pages 1586--1595, 2018.

\bibitem{guo2020dmcp}
Shaopeng Guo, Yujie Wang, Quanquan Li, and Junjie Yan.
\newblock Dmcp: Differentiable markov channel pruning for neural networks.
\newblock In {\em Proceedings of the IEEE/CVF Conference on Computer Vision and
  Pattern Recognition}, pages 1539--1547, 2020.

\bibitem{he2016deep}
Kaiming He, Xiangyu Zhang, Shaoqing Ren, and Jian Sun.
\newblock Deep residual learning for image recognition.
\newblock In {\em Proceedings of the IEEE conference on computer vision and
  pattern recognition}, pages 770--778, 2016.

\bibitem{he2018soft}
Yang He, Guoliang Kang, Xuanyi Dong, Yanwei Fu, and Yi Yang.
\newblock Soft filter pruning for accelerating deep convolutional neural
  networks.
\newblock {\em arXiv preprint arXiv:1808.06866}, 2018.

\bibitem{he2018amc}
Yihui He, Ji Lin, Zhijian Liu, Hanrui Wang, Li-Jia Li, and Song Han.
\newblock Amc: Automl for model compression and acceleration on mobile devices.
\newblock In {\em Proceedings of the European Conference on Computer Vision
  (ECCV)}, pages 784--800, 2018.

\bibitem{howard2017mobilenets}
Andrew~G Howard, Menglong Zhu, Bo Chen, Dmitry Kalenichenko, Weijun Wang,
  Tobias Weyand, Marco Andreetto, and Hartwig Adam.
\newblock Mobilenets: Efficient convolutional neural networks for mobile vision
  applications.
\newblock {\em arXiv preprint arXiv:1704.04861}, 2017.

\bibitem{li2016pruning}
Hao Li, Asim Kadav, Igor Durdanovic, Hanan Samet, and Hans~Peter Graf.
\newblock Pruning filters for efficient convnets.
\newblock {\em arXiv preprint arXiv:1608.08710}, 2016.

\bibitem{liu2018darts}
Hanxiao Liu, Karen Simonyan, and Yiming Yang.
\newblock Darts: Differentiable architecture search.
\newblock {\em arXiv preprint arXiv:1806.09055}, 2018.

\bibitem{liu2017learning}
Zhuang Liu, Jianguo Li, Zhiqiang Shen, Gao Huang, Shoumeng Yan, and Changshui
  Zhang.
\newblock Learning efficient convolutional networks through network slimming.
\newblock In {\em Proceedings of the IEEE International Conference on Computer
  Vision}, pages 2736--2744, 2017.

\bibitem{liu2019metapruning}
Zechun Liu, Haoyuan Mu, Xiangyu Zhang, Zichao Guo, Xin Yang, Kwang-Ting Cheng,
  and Jian Sun.
\newblock Metapruning: Meta learning for automatic neural network channel
  pruning.
\newblock In {\em Proceedings of the IEEE International Conference on Computer
  Vision}, pages 3296--3305, 2019.

\bibitem{liu2018rethinking}
Zhuang Liu, Mingjie Sun, Tinghui Zhou, Gao Huang, and Trevor Darrell.
\newblock Rethinking the value of network pruning.
\newblock {\em arXiv preprint arXiv:1810.05270}, 2018.

\bibitem{lu2020nsganetv2}
Zhichao Lu, Kalyanmoy Deb, Erik Goodman, Wolfgang Banzhaf, and Vishnu~Naresh
  Boddeti.
\newblock Nsganetv2: Evolutionary multi-objective surrogate-assisted neural
  architecture search.
\newblock {\em arXiv preprint arXiv:2007.10396}, 2020.

\bibitem{molchanov2019importance}
Pavlo Molchanov, Arun Mallya, Stephen Tyree, Iuri Frosio, and Jan Kautz.
\newblock Importance estimation for neural network pruning.
\newblock In {\em Proceedings of the IEEE Conference on Computer Vision and
  Pattern Recognition}, pages 11264--11272, 2019.

\bibitem{panda2020nastransfer}
Rameswar Panda, Michele Merler, Mayoore Jaiswal, Hui Wu, Kandan Ramakrishnan,
  Ulrich Finkler, Chun-Fu Chen, Minsik Cho, David Kung, Rogerio Feris, et~al.
\newblock Nastransfer: Analyzing architecture transferability in large scale
  neural architecture search.
\newblock {\em arXiv preprint arXiv:2006.13314}, 2020.

\bibitem{paszke2019pytorch}
Adam Paszke, Sam Gross, Francisco Massa, Adam Lerer, James Bradbury, Gregory
  Chanan, Trevor Killeen, Zeming Lin, Natalia Gimelshein, Luca Antiga, et~al.
\newblock Pytorch: An imperative style, high-performance deep learning library.
\newblock In {\em Advances in neural information processing systems}, pages
  8026--8037, 2019.

\bibitem{radosavovic2020designing}
Ilija Radosavovic, Raj~Prateek Kosaraju, Ross Girshick, Kaiming He, and Piotr
  Dollar.
\newblock Designing network design spaces.
\newblock In {\em Proceedings of the IEEE/CVF Conference on Computer Vision and
  Pattern Recognition}, pages 10428--10436, 2020.

\bibitem{sandler2018mobilenetv2}
Mark Sandler, Andrew Howard, Menglong Zhu, Andrey Zhmoginov, and Liang-Chieh
  Chen.
\newblock Mobilenetv2: Inverted residuals and linear bottlenecks.
\newblock In {\em Proceedings of the IEEE conference on computer vision and
  pattern recognition}, pages 4510--4520, 2018.

\bibitem{tan2019efficientnet}
Mingxing Tan and Quoc~V Le.
\newblock Efficientnet: Rethinking model scaling for convolutional neural
  networks.
\newblock {\em arXiv preprint arXiv:1905.11946}, 2019.

\bibitem{wong2018transfer}
Catherine Wong, Neil Houlsby, Yifeng Lu, and Andrea Gesmundo.
\newblock Transfer learning with neural automl.
\newblock In {\em Advances in Neural Information Processing Systems}, pages
  8356--8365, 2018.

\bibitem{yu2019autoslim}
Jiahui Yu and Thomas Huang.
\newblock Autoslim: Towards one-shot architecture search for channel numbers.
\newblock {\em arXiv preprint arXiv:1903.11728}, 2019.

\bibitem{zhou2020econas}
Dongzhan Zhou, Xinchi Zhou, Wenwei Zhang, Chen~Change Loy, Shuai Yi, Xuesen
  Zhang, and Wanli Ouyang.
\newblock Econas: Finding proxies for economical neural architecture search.
\newblock In {\em Proceedings of the IEEE/CVF Conference on Computer Vision and
  Pattern Recognition}, pages 11396--11404, 2020.

\bibitem{Zoph_2018_CVPR}
Barret Zoph, Vijay Vasudevan, Jonathon Shlens, and Quoc~V. Le.
\newblock Learning transferable architectures for scalable image recognition.
\newblock In {\em Proceedings of the IEEE Conference on Computer Vision and
  Pattern Recognition (CVPR)}, June 2018.

\end{thebibliography}
}

\clearpage
\appendix

\section{Training hyperparameters}\label{app:hyperparam}
We used PyTorch~\cite{paszke2019pytorch} as our deep learning framework. We largely follow~\cite{radosavovic2020designing} for training hyperparameters. Specifically, learning rate grows linearly from 0 to $0.2s$ within the first 5 epochs from 0 where $s$ depend on batch size $B$, \textit{i.e.}, $s=\frac{B}{256}$. We use batch size of 1024 and distributed training over 8 GPUs and we have not used synchronized batch normalization layers. We set the training epochs to be 100. For optimizers, we use stochastic gradient descent (SGD) with 0.9 Nesterov momentum. As for data augmentation, we have adopted 0.1 label smoothing, random resize crop, random horizontal flops, and RandAugment~\cite{cubuk2020randaugment} with parameter $N=2$ and $M=9$ following common practice in popular repository\footnote{\url{https://github.com/rwightman/pytorch-image-models} (We use their implementation for RandAugment and use `rand-m9-mstd0.5' as the value for the `aa' flag.}. Note that these training hyperparameters are fixed for all experiments. 

For hyperparameters specific to width optimization algorithms, we largely follow the hyperparameters used in respective methods. Specifically, we use 40 epochs to search for optimized width for all three algorithms. We enlarge the network by $1.5\times$ for DMCP and AutoSlim. Since MorphNet has FLOPs-aware regualrization, we normalize the FLOPs for each network and use $\lambda=1$ for all experiments.

\section{Similarity among width multipliers}
In Section~\ref{sec:exp}, we have analyzed the similarity between $\vw^*$ and $\hat{\vw}^*$ in the accuracy space. Here, we show that $\vw^*$ and $\hat{\vw}^*$ are in fact similar in the vector space using cosine similarity.

\begin{figure*}[t!]
     \centering
     \begin{subfigure}[b]{0.33\textwidth}
         \centering
         \includegraphics[width=\textwidth]{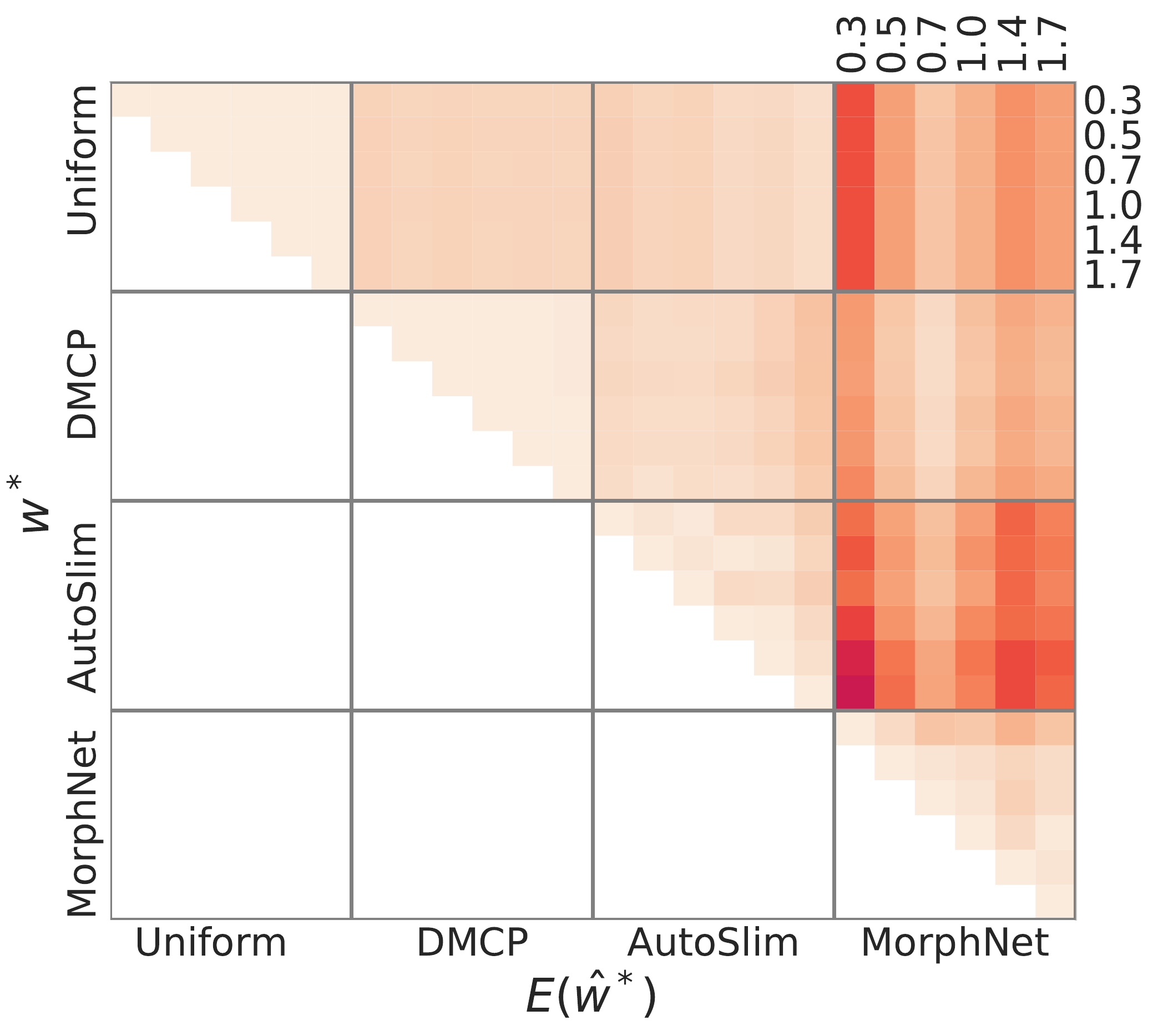}
         \caption{ResNet18, width projection}
     \end{subfigure}
     \hfill
     \begin{subfigure}[b]{0.33\textwidth}
         \centering
         \includegraphics[width=\textwidth]{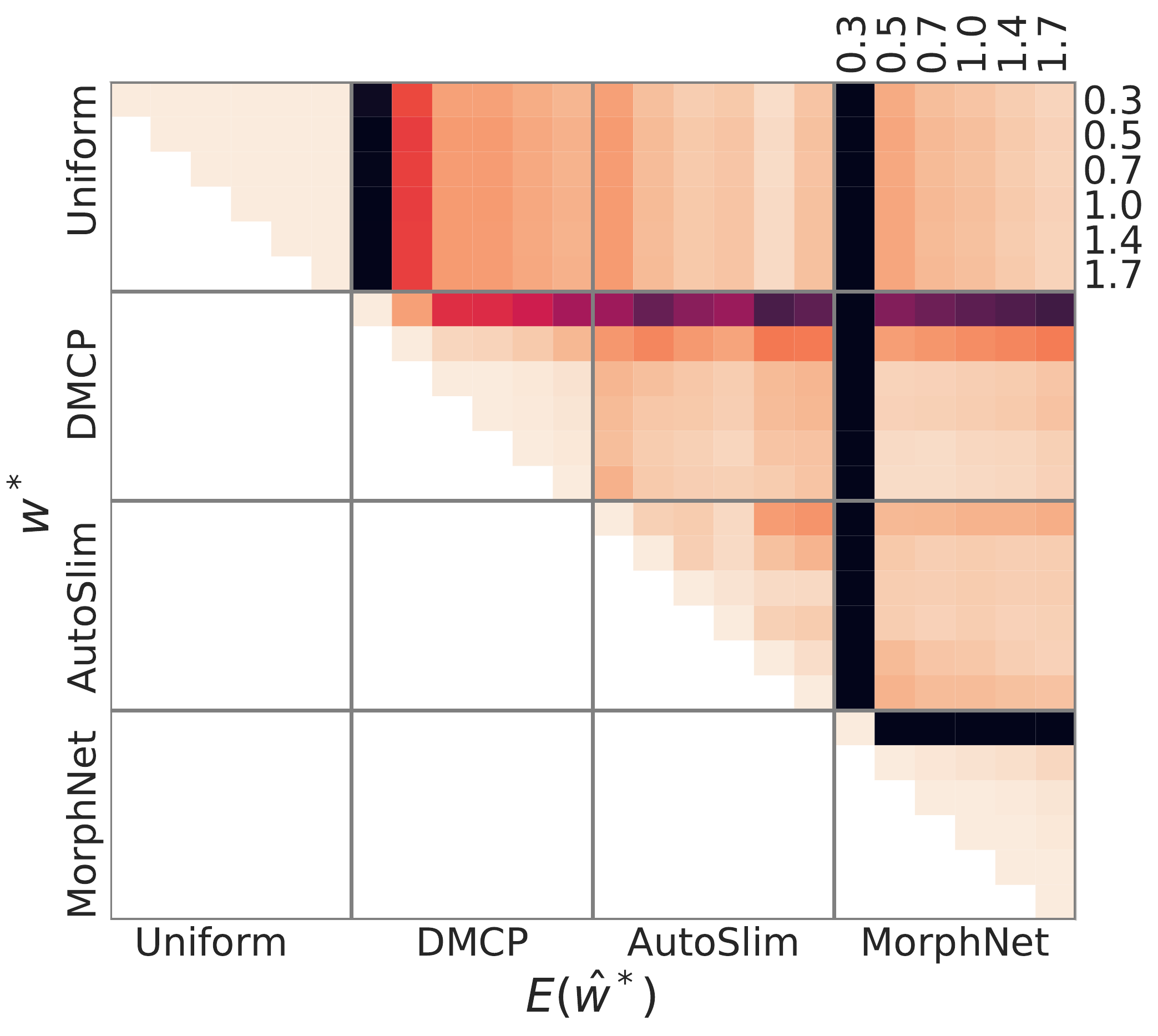}
         \caption{MobileNetV2, width projection}
     \end{subfigure}
     \hfill
     \includegraphics[width=0.06\textwidth]{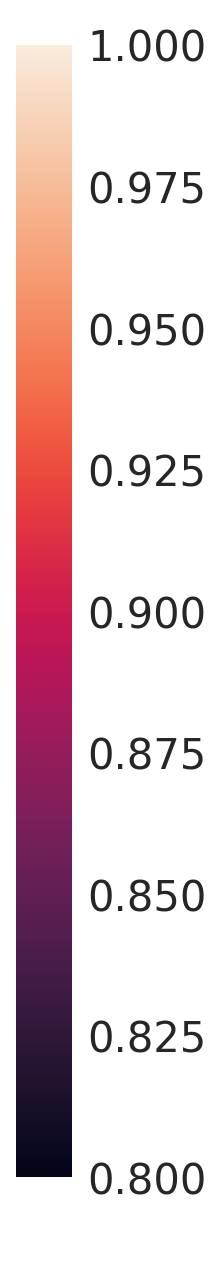}
     \\
     \begin{subfigure}[b]{0.33\textwidth}
         \centering
         \includegraphics[width=\textwidth]{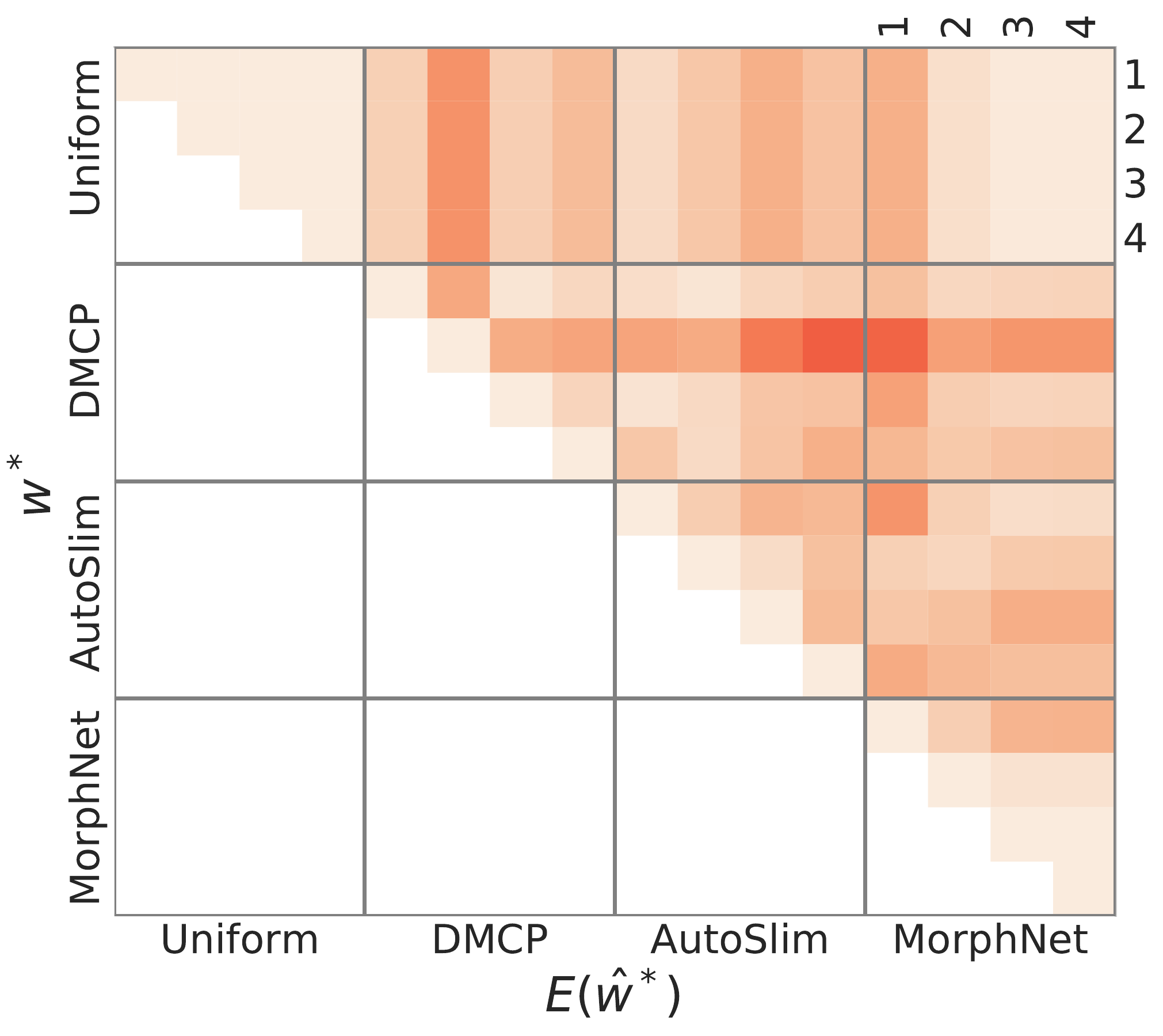}
         \caption{ResNet18, depth projection}
     \end{subfigure}
     \hfill
     \begin{subfigure}[b]{0.33\textwidth}
         \centering
         \includegraphics[width=\textwidth]{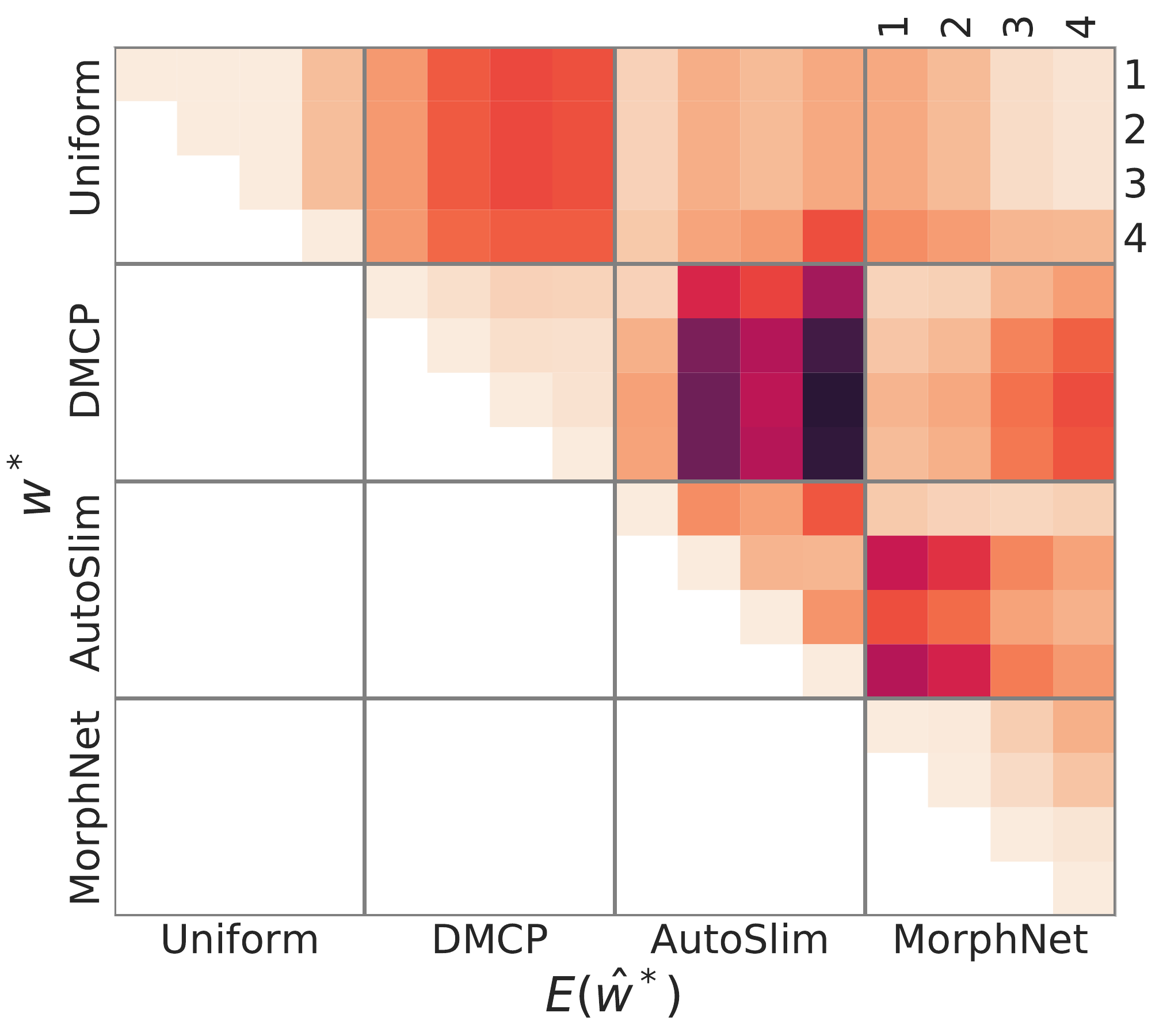}
         \caption{MobileNetV2, depth projection}
     \end{subfigure}
     \hfill
     \includegraphics[width=0.06\textwidth]{figs/colorbar.png}
     \\
     \begin{subfigure}[b]{0.33\textwidth}
         \centering
         \includegraphics[width=\textwidth]{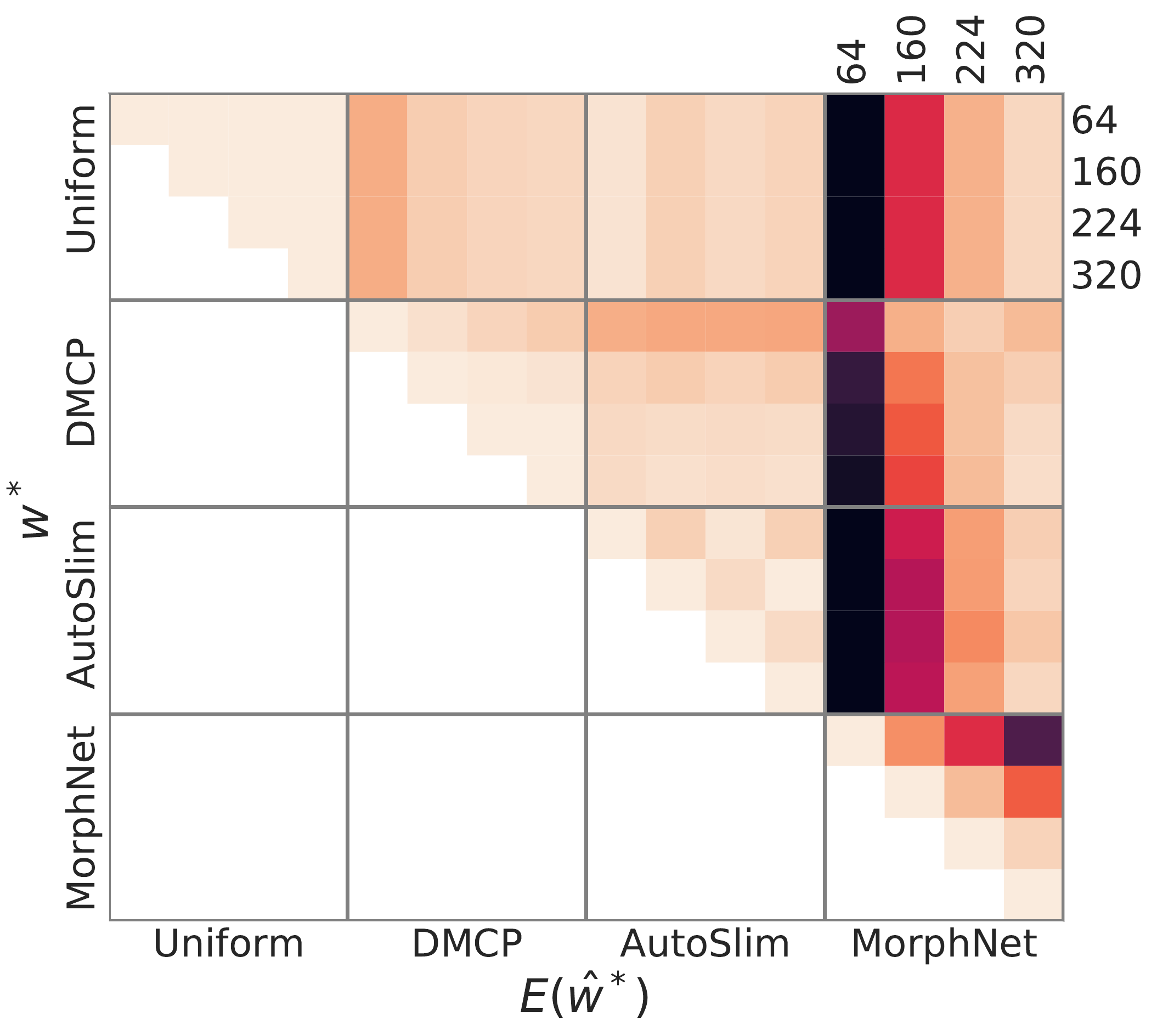}
         \caption{ResNet18, resolution projection}
         \label{fig:resnet18-res-sim}
     \end{subfigure}
     \hfill
     \begin{subfigure}[b]{0.33\textwidth}
         \centering
         \includegraphics[width=\textwidth]{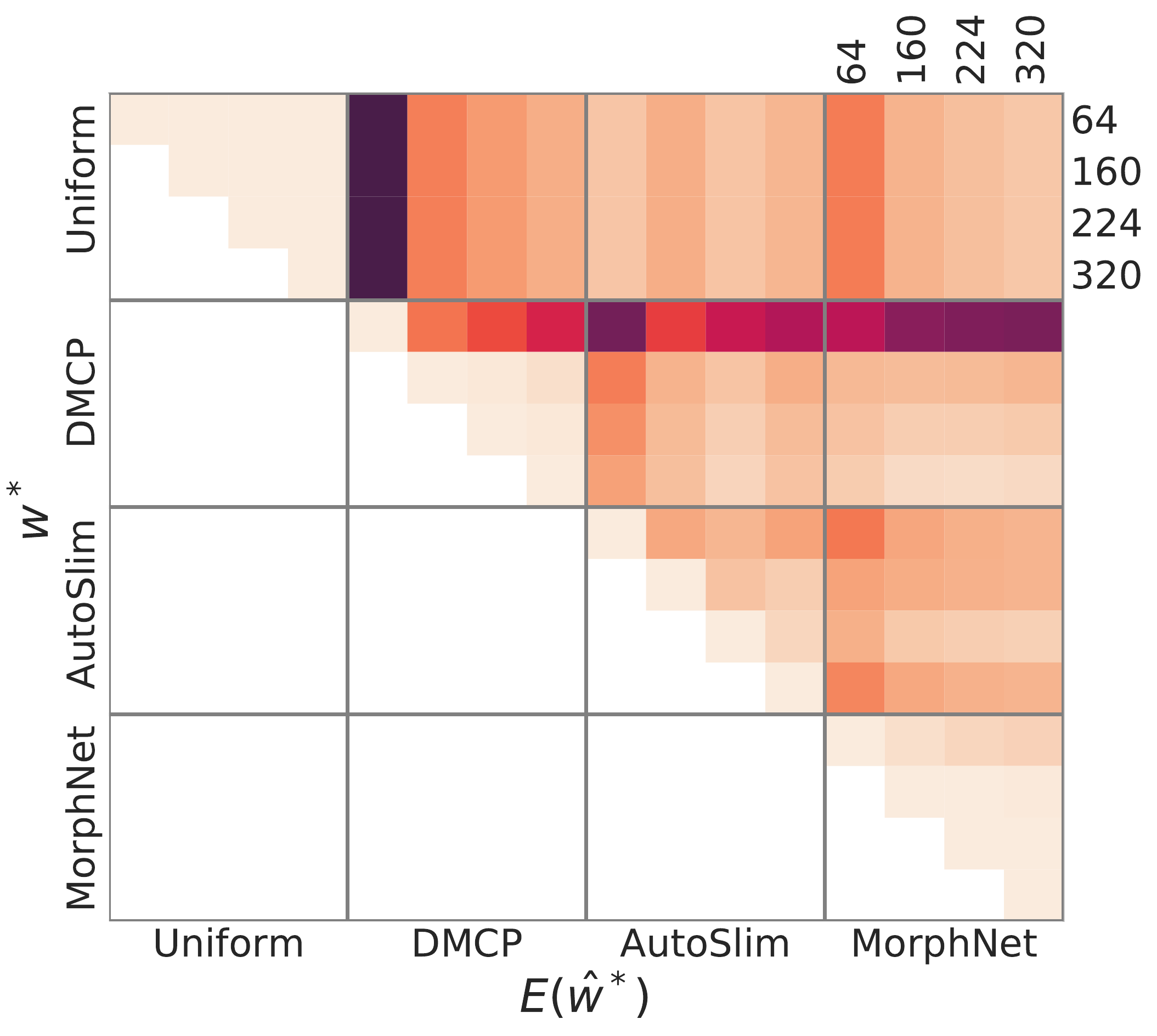}
         \caption{MobileNetV2, resolution projection}
         \label{fig:mbv2-res-sim}
     \end{subfigure}
     \hfill
     \includegraphics[width=0.06\textwidth]{figs/colorbar.png}
     \\
     \begin{subfigure}[b]{0.33\textwidth}
         \centering
         \includegraphics[width=\textwidth]{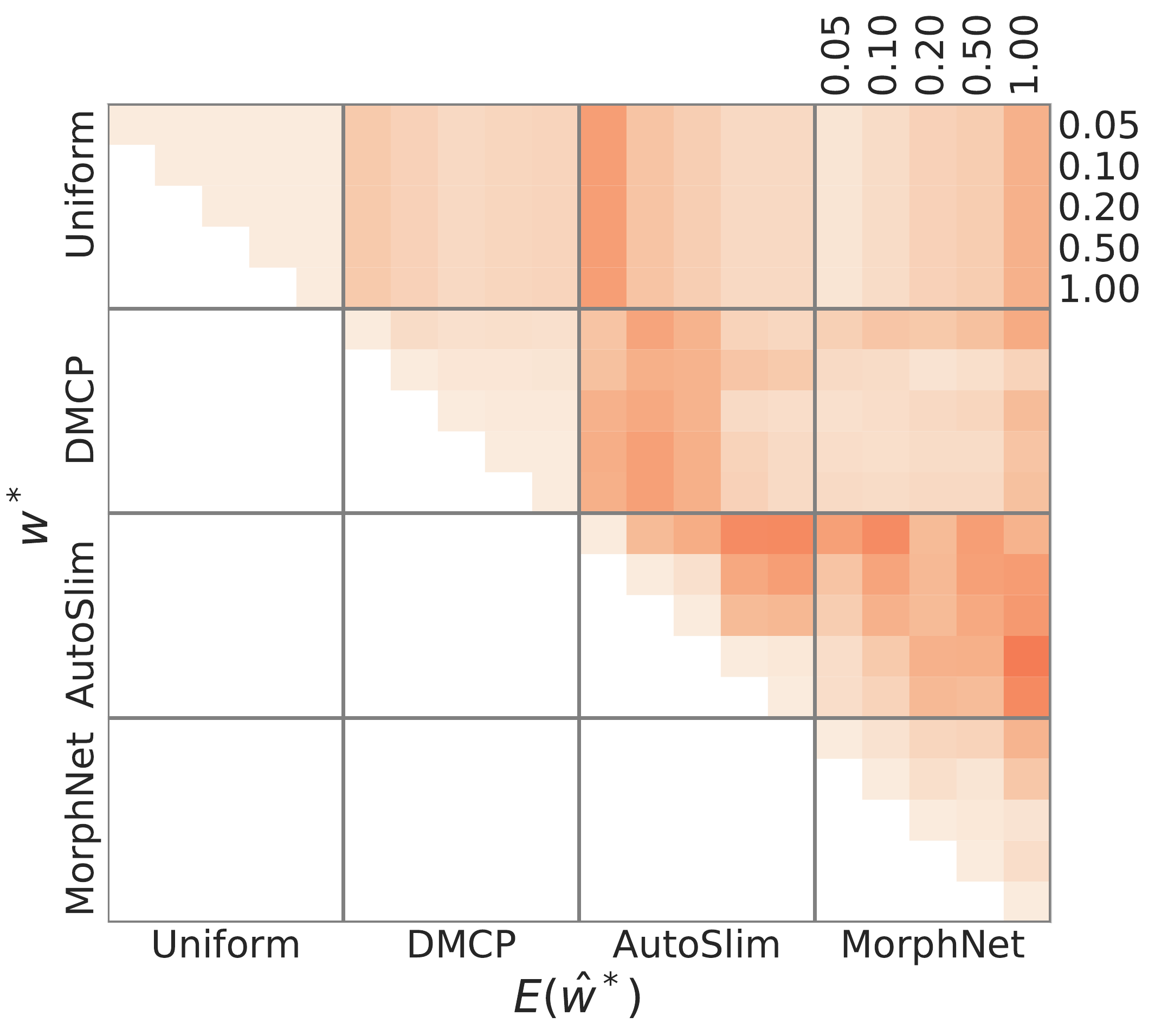}
         \caption{ResNet18, dataset size projection}
     \end{subfigure}
     \hfill
     \begin{subfigure}[b]{0.33\textwidth}
         \centering
         \includegraphics[width=\textwidth]{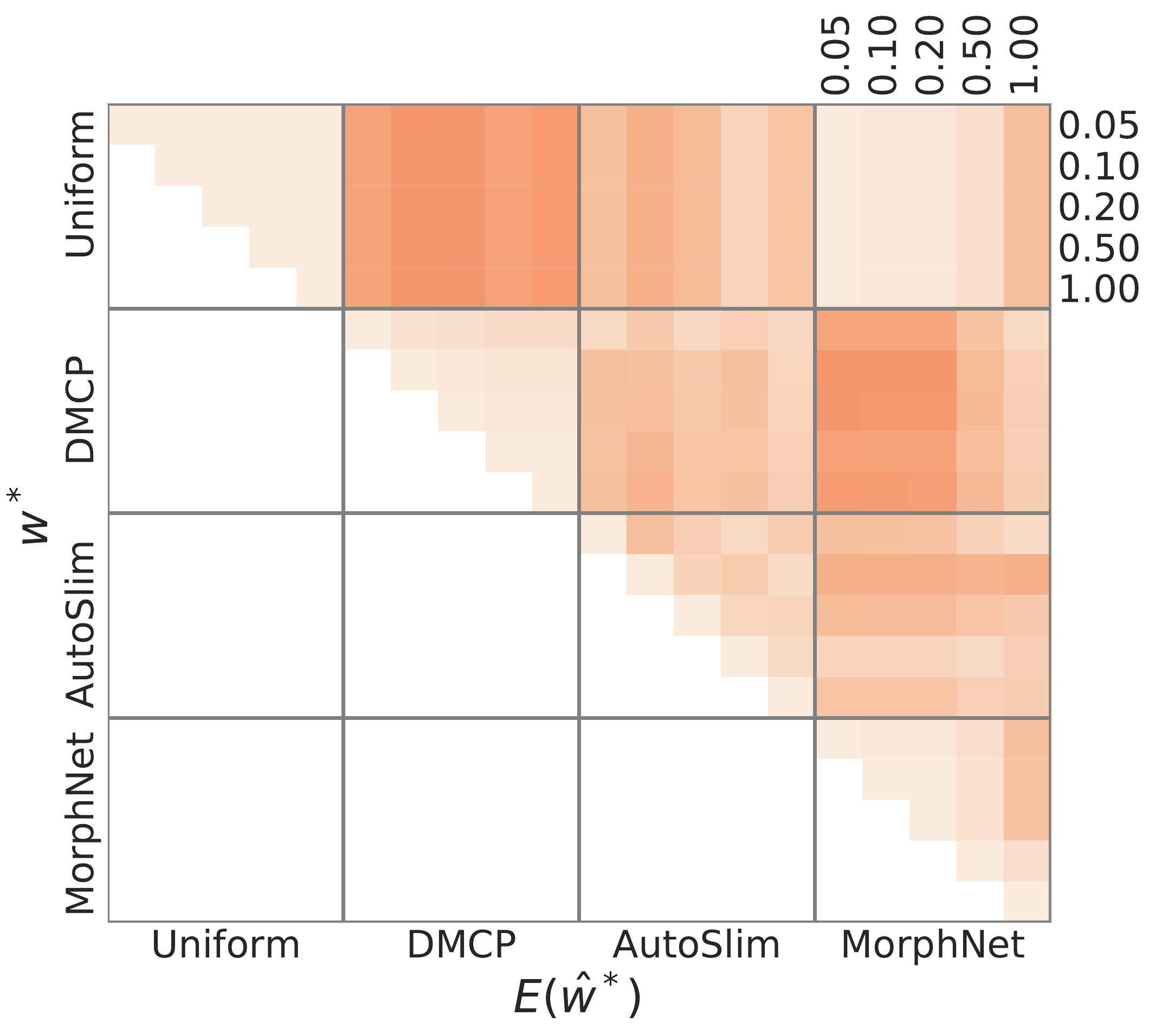}
         \caption{MobileNetV2, dataset size projection}
     \end{subfigure}
     \hfill
     \includegraphics[width=0.06\textwidth]{figs/colorbar.png}
    \caption{Pairwise cosine similarity between $\vw^*$ and $E(\hat{\vw}^*)$ for different width optimization algorithms and projection strategies. Within each methods (diagonal blocks), $\vw^*$ and $E(\hat{\vw}^*)$ are generally similar.}
    \label{fig:sim}
\end{figure*}

\end{document}